\pdfoutput=1
\documentclass[10pt,twocolumn,letterpaper]{article}

\usepackage[pagenumbers]{cvpr} 

\usepackage{times}  
\usepackage{helvet}  
\usepackage{courier}  
\usepackage[hyphens]{url}  
\usepackage{graphicx} 
\usepackage{natbib}  
\usepackage{caption} 
\usepackage{algorithm}
\usepackage{algorithmic}

\usepackage{amsmath}
\usepackage{amssymb}
\usepackage{tcolorbox}
\usepackage{dsfont}
\usepackage{multirow}
\usepackage{xcolor}
\usepackage{colortbl}
\usepackage{rotating}
\usepackage{booktabs}
\usepackage{tabularray}
\usepackage{subcaption} 
\usepackage{pifont}
\usepackage{makecell}
\usepackage{float}

\DeclareMathOperator*{\argmin}{arg\,min}
\DeclareMathOperator*{\argmax}{arg\,max}

\newcommand{\norm}[1]{\left\lVert#1\right\rVert}

\definecolor{lightblue}{RGB}{170, 216, 230} 
\definecolor{cvprblue}{rgb}{0.21,0.49,0.74}
\usepackage[pagebackref,breaklinks,colorlinks,allcolors=cvprblue]{hyperref}

\def\paperID{7912} 
\def\confName{CVPR}
\def\confYear{2025}

\title{Can't Slow me Down: Learning Robust and Hardware-Adaptive Object Detectors against Latency Attacks for Edge Devices}

\author{\text{Tianyi Wang, Zichen Wang, Cong Wang\thanks{Corresponding author}, Yuanchao Shu, Ruilong Deng, Peng Cheng, Jiming Chen}\\
Zhejiang University, Hangzhou, China, \\
{\tt\small \{wty1998, withnorman, cwang85, ycshu, dengruilong, lunarheart, cjm\}@zju.edu.cn}\\
}

\begin{document}
\maketitle
\begin{abstract}
Object detection is a fundamental enabler for many real-time downstream applications such as autonomous driving, augmented reality and supply chain management. However, the algorithmic backbone of neural networks is brittle to imperceptible perturbations in the system inputs, which were generally known as misclassifying attacks. By targeting the real-time processing capability, a new class of latency attacks are reported recently. They exploit new attack surfaces in object detectors by creating a computational bottleneck in the post-processing module, that leads to cascading failure and puts the real-time downstream tasks at risks. In this work, we take an initial attempt to defend against this attack via background-attentive adversarial training that is also cognizant of the underlying hardware capabilities. We first draw system-level connections between latency attack and hardware capacity across heterogeneous GPU devices. Based on the particular adversarial behaviors, we utilize objectness loss as a proxy and build background attention into the adversarial training pipeline, and achieve a reasonable balance between clean and robust accuracy. The extensive experiments demonstrate the defense effectiveness of restoring real-time processing capability from $13$ FPS to $43$ FPS on Jetson Orin NX, with a better trade-off between the clean and robust accuracy. The source code is available at \url{https://github.com/Hill-Wu-1998/underload}
\end{abstract}    
\section{Introduction}
Real-time object detection lies at the heart of a wide range of downstream applications such as autonomous driving~\cite{liang2024aide}, drone navigation~\cite{wang2023generalized} and video surveillance~\cite{lu23aaai,jiang21mobicom}. From the early detectors such as (Faster) RCNN~\cite{rcnn,fastercnn} to the latest versions of the YOLO family~\cite{yolov1,yolov5,yolov8}, we have seen incredible achievements of performance-efficiency improvements, \eg, reaching 56-60\% mAP on the MS-COCO benchmark with more than 30 FPS on embedded NVIDIA Jetson boards.

\begin{figure}[t]
\centering
\subfloat[Original image]{\includegraphics[width=0.31\columnwidth, height=6cm]{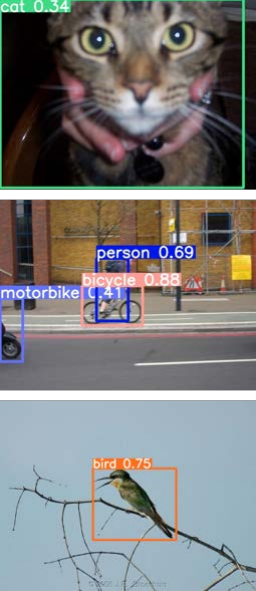}\label{fig:1.1.a}}\hspace{0.5pt}
\subfloat[Latency attack]{\includegraphics[width=0.31\columnwidth, height=6cm]{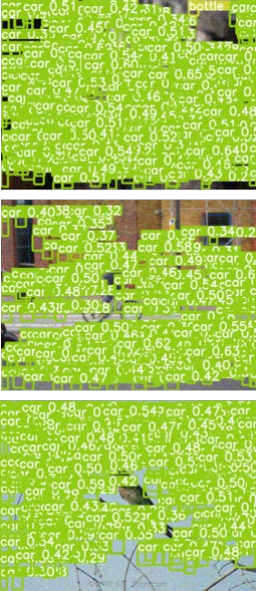}\label{fig:1.1.b}}\hspace{0.5pt}
\subfloat[Our defense]{\includegraphics[width=0.31\columnwidth, height=6cm]{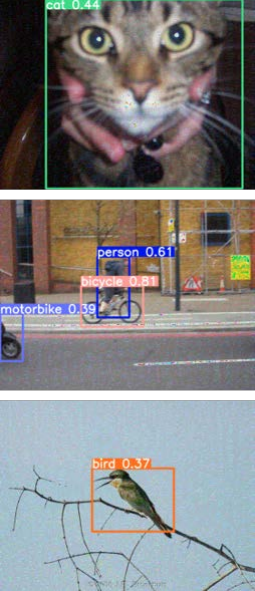}\label{fig:1.1.c}} 
\vspace{-0.05in}
\caption{\small{Visualization of \emph{latency attacks}~\cite{ps} generating an overwhelming number of ``phantom objects'' and the effectiveness of our defense.}}
\label{fig1.1:adv_demo}
\vspace{-0.15in}
\end{figure}

However, in the shadow of these tremendous successes, a new threat called \emph{latency attack} lurks to impair the real-time processing capability of object detectors~\cite{chen2024overload,ma2024slowtrack,shumailov2021sponge,wang2021daedalus}, which employ hand-crafted Non-Maximum Suppression (NMS) to eliminate duplicate objects in the post-processing stage. By targeting the computational bottleneck inside the NMS module~\cite{nms_objdet}, these attacks create more than thousands ``phantom objects'' with gradient-based adversarial techniques to congest the NMS processing pipeline as shown in Fig. \ref{fig1.1:adv_demo}. In addition to the known attacks against object detectors such as disappearing, mis-classifying and mis-location~\cite{tog, appering_attack, advattack}, it expands the attacker's weapon arsenal with high risks of jeopardizing real-time applications on edge systems. \Eg, any latency in detecting an obstacle in autonomous driving could lead to cascading failure in the sensor fusion, decision and steering control subsystems.

Unfortunately, with the arms race on the attack side, we have seen a paucity of research to defend against these new vulnerabilities. A quick fix is to set hard limits on the number  objects~\cite{chen2024overload}, whereas such limit is hard to select for different applications since a large value would still allow phantom objects to pass and a small one might falsely reject correct instances in case a large number of objects are present, especially in crowd counting and traffic monitoring applications. A comprehensive solution is to completely eliminate the hand-crafted NMS~\cite{dert,deform_dert}. However, this requires to change all the legacy software dependencies on the edge firmware and NMS-free architectures are still far from achieving real-time performance on edge devices~\cite{detr_beats_yolo}. Admittedly, the computational bottleneck in NMS is not straightforward to circumvent: our system-level analysis unveils that it not only comes from limited computational power (on edge devices), but also extensive data transfer between GPU-CPU that is not discussed in the previous works~\cite{chen2024overload,wang2021daedalus,ps}. Would heterogeneous GPUs have different capacities under these attacks -- are high-end GPUs free from such attack? If not and the bottleneck persists, without removing the NMS module, how can we thwart latency attacks? By giving a performance requirement, can we guarantee such requirement by learning robust, adaptive object detectors on different hardware accelerators? 

To answer these questions, in this paper, we propose \texttt{Underload} against \texttt{Overload}~\cite{chen2024overload} and a series of latency attacks~\cite{ma2024slowtrack,wang2021daedalus, ps} via a hardware-adaptive, background-attentive Adversarial Training (AT) mechanism. We first leverage system-level analysis to identify processing bottlenecks and orchestrate these findings to model the GPU capacity with the number of candidate bounding boxes. We discover the usage of objectness loss can be served as a proxy to eliminate phantom objects. By delving into the unique adversarial behaviors of latency attacks, we also find that the background region typically consists of non-robust features~\cite{kim2021distilling}, that are more vulnerable. To this end, we design background-attentive AT mechanism, which weighs more on the background semantics to achieve a better balance between the robust and clean accuracy. The main contributions are summarized below.

\begin{itemize}
    \item[\ding{172}] \textbf{Motivation.} Based on extensive programming analysis and system profiling, we discover interesting phenomenons on the migration of processing bottlenecks from compute-bound to memory-bound operations between edge and desktop GPUs. These findings allow us to establish a bridge between attack strength and heterogeneous hardware capacity with a deeper understanding of the system impact on various types of GPUs.
    
    \item[\ding{173}] \textbf{Methodology.} We perform in-depth analysis of the objectness loss and draw connection with the latency attacks. Then we find unique footprint regarding the discriminative boundary margins between the background and object regions. Our defense successfully designs these factors into the AT pipeline and achieves a reasonable balance between clean and robust accuracy by exploiting the robust/non-robust features.    
    
    \item[\ding{174}] \textbf{Evaluation}. We perform extensive experiments across the widely-adopted YOLOv3, YOLOv5 and the latest YOLOv8 (anchor-free) on embedded GPUs (Jetson Xavier/Orin NX), desktop GPUs (4070Ti Super) and multi-tenant cloud GPUs (A100). The results show that our approach achieves up to 8-10\% gain on robust accuracy compared to previous existing defense of MTD~\cite{mtd} and OOD~\cite{objadv} with less clean accuracy loss. Under various latency attacks, our defense is also able to restore the real-time processing capability from $13$ FPS to $43$ FPS on Jetson Orin NX as well as different types of GPUs.
    
\end{itemize}
\section{Background and Related Works}  \label{sec:related_work}

\textbf{Object Detection} includes both classification and precise localization of objects within a digital image, addressing the question of ``what'' and ``where'' the objects are~\cite {obj_det_survey}. The types of object detectors mainly include CNN-based one-stage detectors~\cite{ssd,yolov1,yolov3,yolov5,yolov8}, two-stage detectors~\cite{rcnn, fastercnn,zhang2020dynamic}, transformer-based detectors such as DETR~\cite{dert,deform_dert,dai2021dynamic} and diffusion detectors~\cite{chen2023diffusiondet}. Two-stage detectors first involve region proposal with the objects of interest, then the second stage classifies and refines the bounding boxes for more precise localization~\cite{rcnn,fastercnn,zhang2020dynamic}. On the other hand, one-stage detectors transform object detection into a regression problem that directly obtains classification and localization with a single pass, without the extensive RoI extraction~\cite{yolov1, yolov3, yolov5, ssd}. Hence, the research community has embraced one-stage detectors due to their simplicity, fast response and hardware affinity across a wide variety of embedded devices. As a direct response to the latency attacks against the YOLO family, this paper focuses on the vulnerabilities in one-stage detectors.

\textbf{Non-Maximum Suppression.} NMS serves as an essential post-processing backend and becomes an indispensable step for different object detectors~\cite{yolov5,ssd,fastercnn}. The main purpose is to consolidate and remove redundant objects in each object cluster and return a final bounding box identified as the local maxima, with peaks exceeding their neighboring values by excluding the maxima themselves~\cite{nms}. A different variety of NMS methods have been proposed such as GIoU~\cite{giou}, CIoU~\cite{zheng2021enhancing}, $\alpha$-IoU~\cite{he2021alpha} and Wise-IoU~\cite{tong2023wise}. However, as per to both prior and our analysis, NMS poses inherent security vulnerabilities and opens up a unique attack surface to latency attacks due to its programming structure and system implementation described in the next section.

\subsection{Latency Attacks} 

The rationale behind latency attacks originates from the classic \emph{Denial of Service} attacks that originally cause congestion in networking services. Sponge example is the first attack in AI systems that aims to increase energy consumption and inference time by firing up more activations in NLP models that lead to more multiply-add operations~\cite{shumailov2021sponge}. Surprisingly, its impact is less significant in vision tasks when the default activation values are already non-zero for most images. This makes the follow-up works to turn their targets to the hand-crafted NMS module~\cite{wang2021daedalus,ps,chen2024overload,ma2024slowtrack}. 

Daedalus is the pioneering work that exploits the vulnerability of NMS~\cite{wang2021daedalus}. Three adversarial losses are developed to make the final outputs contain an extremely high density of false positives. Phantom Sponge enhances Daedalus's methodology by training a universal adversarial perturbation, and analyzes the NMS execution time to formulate the latency attacks against object detection models. Recent efforts extend these attacks towards resource-constrained edge devices~\cite{chen2024overload} and autonomous driving backends~\cite{ma2024slowtrack}. Overload analyzes the NMS complexity, which analytically confirms latency extension by increasing candidate bounding boxes entering the NMS~\cite{chen2024overload}. Furthermore, it augments the attack efficacy by incorporating spatial attention. SlowTrack adopts latency attacks in the autonomous driving scenario and raises the vehicle crash rate to 95\% by executing latency attacks on camera-based autonomous driving systems~\cite{ma2024slowtrack}. There are also another types of latency attacks that does not target object detectors~\cite{ilfo, nmtsloth}. All relevant works above remain on the offensive side to exploit NMS as a new attack surface. To our best knowledge, this is the first attempt that builds specialized defense against latency attacks for object detectors running on millions of edge devices. 

\section{Understanding Latency Attacks}

\subsection{Algorithmic Vulnerability from NMS}

\begin{figure}[t]
\centering
\includegraphics[width=0.98\columnwidth]{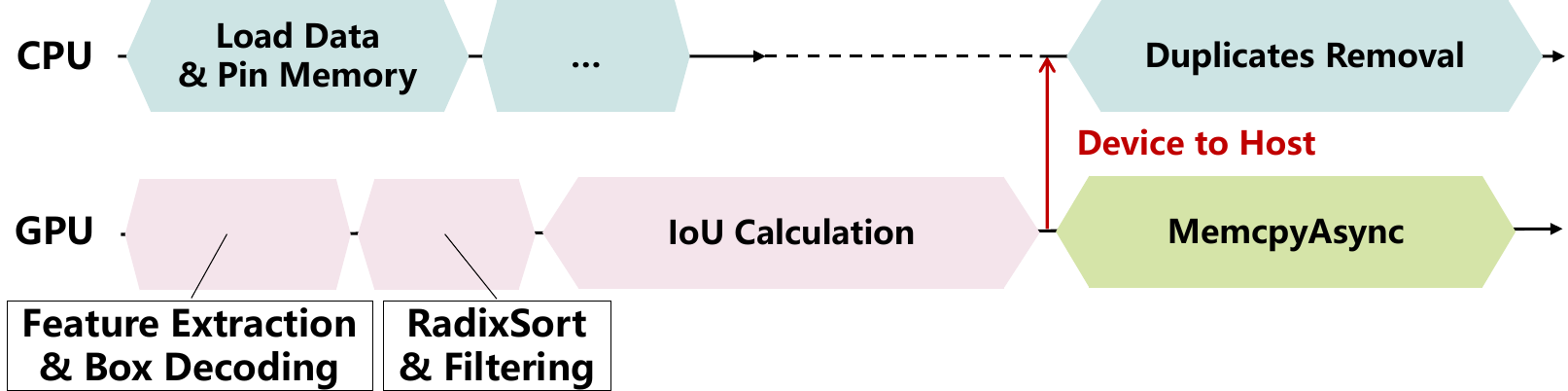}
\caption{\small{Interactions between GPU-CPU in the NMS.}}
\label{fig:nms_demo}
\end{figure}

After the feature extractor, the model outputs candidate bounding boxes defined by key attributes of height, width, location, confidence score, and categorical probabilities~\cite{ps}. Then NMS removes duplicated boxes by consolidating them with closely-matched positions into a single one, i.e., boxes with an \emph{intersection over union} (IoU) value over a specified threshold are preserved, and IoU quantifies the overlap between two boxes relative to their union~\cite{wang2021daedalus}. The procedures of NMS are detailed in Algorithm~\ref{alg:nms} and the interactions between the GPU-CPU are visualized in Fig. \ref{fig:nms_demo}. Specifically, Ln $3$-$5$ perform filtering, radixsort and pairwise IoU calculation of candidate boxes (on GPU); Ln $7$-$8$ remove unqualified boxes by comparing with the threshold (on CPU). For each candidate box, data transfer between the GPU and CPU is required. 

\tcbset{logic/.style={colframe=lightgray, colback=white, arc=3mm, left=-4pt, right=0pt, boxrule=0.5pt}}
\tcbset{compute/.style={colframe=black, colback=white, arc=3mm, left=-4pt, right=0pt, boxrule=0.5pt}}

\begin{algorithm}
\footnotesize
\caption{Non-Maximum Suppression (NMS)}
\label{alg:nms}
\textbf{Input}: $\mathcal{B}$: candidate boxes, $\mathcal{S}$: scores, $\Omega_{nms}$: NMS threshold. \\
\textbf{Output}: $\mathcal{R}$: NMS result, $\mathcal{S}$: updated scores after NMS. \\
\begin{algorithmic}[1]
\vspace{-0.12in}
\WHILE{$B \neq \emptyset$}
    \vspace{-0.05in}
    \begin{tcolorbox}[compute]
    \vspace{-0.08in}
    \STATE ~~~~\textcolor{blue}{/* Operations on the GPU */}
    \STATE ~~$\mathcal{M} \leftarrow \arg\max \mathcal{S}$, $\mathcal{R} \leftarrow \mathcal{R} \cup \{M\}$, $\mathcal{M} \leftarrow \mathcal{M}/\mathcal{B}$.
    \STATE ~~\textbf{for}~each $b_i \in \mathcal{B}$~\textbf{do}
        \STATE ~~~~ $IoU_i = IoU(M, b_i)$.
        \vspace{-0.1in}
        \end{tcolorbox}
\begin{tcolorbox}[logic]
        \vspace{-0.1in}
        \STATE ~~~~\textcolor{red}{/* Operations on the CPU */}
        \STATE ~~~~ \textbf{if}~$IoU_i > \Omega_{nms}$~\textbf{then}
        \vspace{-0.03in}
            \STATE ~~~~~~~ $\mathcal{B} \leftarrow \mathcal{B}/b_i$, $\mathcal{S} \leftarrow \mathcal{S}/S_{b_i}$. 
            \vspace{-0.1in}
            \end{tcolorbox}
        \STATE ~~~~\textbf{end if}
    \STATE ~~\textbf{end for}
\ENDWHILE
\STATE \textbf{return} $\mathcal{R}$, $\mathcal{S}$
\end{algorithmic}
\end{algorithm}

\noindent \textbf{Vulnerability (Time Complexity).} The execution time of NMS increases quadratically when the number of candidate bounding boxes exceeds a threshold $N_t$.

\begin{equation}
\small
T_{\text{nms}}=\left\{
\begin{aligned}
\label{equ:num_time}
T_{\text{base}} ,& ~~~|C| \leq N_t \\
a |C|^2 ,& ~~~|C| > N_t \\
\end{aligned}
\right.
\end{equation}

This vulnerability is exploited by~\cite{chen2024overload} as an analytical basis for latency attacks. Shown in Eq.~\eqref{equ:num_time}, the processing time remains a constant $T_{\text{base}}$ if the box count $|C|$ is below the threshold $N_t$, but increasing quadratically when $|C|$ is larger than $N_t$ with a magnifying factor $a$. By utilizing adversarial perturbations, the attacker's objective is to maximize the box confidence before NMS, thereby feeding it with more candidate boxes. The adversarial objective can be summarized as,
\begin{equation}
\small
\mathcal{L}_{\text{adv}} = \underbrace{\underbrace{\max~{\mathcal{L}_{\text{conf}}}}_{\texttt{Overload}~\text{\cite{chen2024overload}}} + \rho \min(\mathcal{L}_{\text{bbox}} + \mathcal{L}_{\text{max IoU}})}_{\texttt{PhantomSponge}\text{\cite{ps}}}, \label{equ:attack_loss_func} \\
\end{equation}
where \texttt{Overload} maximizes the box confidence~\cite{chen2024overload} and \texttt{Phantom Sponge} utilizes auxiliary losses $\mathcal{L}_{\text{bbox}}$ to reduce the area of bounding boxes for lower overall IoU and $\mathcal{L}_{\text{max IoU}}$ to preserve the detection of original objects~\cite{ps}. The discussions above only provide algorithmic analysis, to unveil the impact on the computer architecture and system level, we provide a deeper understanding of how latency attacks functionalize across heterogeneous GPUs.  

\subsection{System-level Impacts of Latency Attacks}

\begin{figure}[t]
\vspace{-0.1in}
\centering
\subfloat[\footnotesize{NMS processing time}]{
\includegraphics[width=0.49\columnwidth]{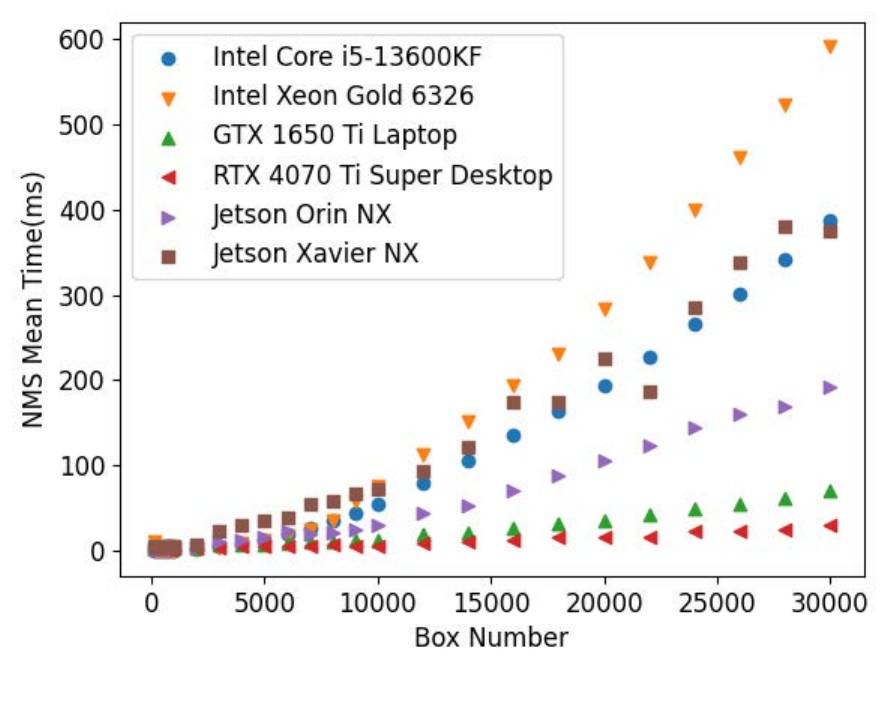}
\label{fig:3.1.a}
}
\subfloat[\footnotesize{\% of different operations}]{
\includegraphics[width=0.49\columnwidth]{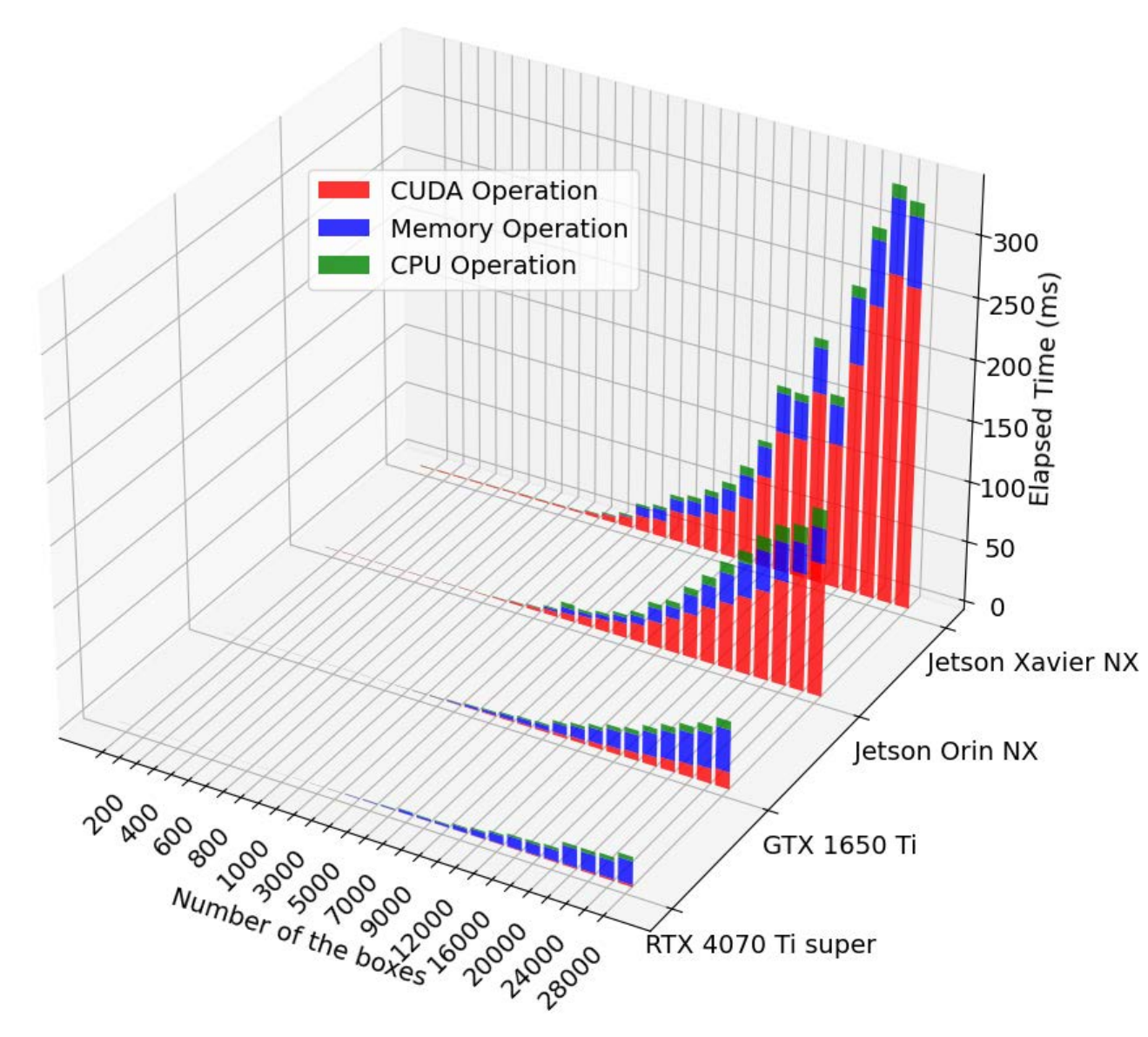}
\label{fig:3.1.b}
}
\vspace{-0.05in}
\caption{\small{System effects of latency attacks. a) \# objects vs. NMS processing time on heterogeneous computing devices; b) percentage of GPU CUDA, memory transfer and CPU logic operations on embedded and desktop GPUs.}}
\label{fig3.1:attack_result}
\vspace{-0.15in}
\end{figure}

Our goal is to differentiate the adversarial impacts on GPUs with heterogeneous processing power as profiled in Fig.~\ref{fig3.1:attack_result} with two interesting insights below.

\noindent \textbf{Observation 1 (Compute-Bound on Embedded GPUs).} On resource-constrained edge devices that lack GPU cores, latency attacks make real-time processing \emph{compute-bound}. It suggests that edge devices need to increase the number of GPU cores to tolerate such attacks.

This is justified by Fig. \ref{fig:3.1.b} that both Jetson devices have parabolic increase of CUDA operations with an increasing number of bounding boxes (384 vs. 1024 CUDA cores of Xavier and Orin NX, respectively). It is also cross-validated in Fig. \ref{fig:3.1.a} that both Intel CPUs rank the lowest if NMS is executed on the CPUs with limited logical cores.

\noindent \textbf{Observation 2 (Memory-Bound on Desktop GPUs).} As the number of GPU cores increases, the bottleneck is mitigated but not vanished, but gradually migrating from computation to data transfer between the GPU and CPU due to logical operations, which suggests that one should increase memory bandwidth and parallelization of CPU cores to eliminate latency attacks. The memory bottleneck becomes more prominent on multi-tenant cloud platforms since the memory bandwidth is shared and interference from users on the same physical machine would magnify the attack impact, which is validated by our experiments using A100 GPUs in Sec. \ref{sec:evaluation}. 

The observation above is evidenced from Fig. \ref{fig:3.1.b} that both 1650Ti and 4070Ti GPUs are dominated by memory operations between the GPU and CPU to retain the boxes after IoU calculation. Based on these results, the number of candidate boxes into the NMS module $\mathbb{E}[f_\theta(x)]_{\text{box}}$ should be bounded by the capacity of the hardware accelerators to satisfy the real-time application requirement. Given the response time $T$, \eg, $T=33.3$ ms for standard rate at 30 FPS and $T=16.6$ ms for high-end tasks at 60 FPS.

\begin{equation}
\small
T_{\text{nms}}  = (\alpha \frac{|C|^2}{S_\text{IoU}} + \beta \frac{|C|}{B}) < T - T_{\text{backbone}}  \\
\end{equation}
\begin{equation}
\small
\mathbb{E}[f_\theta(x)]_{\text{box}} < \frac{S_\text{IoU}}{2 \alpha}\sqrt{\frac{\beta^2}{B^2}-4 \frac{\alpha}{S_\text{IoU}} (T - T_{\text{backbone}})}      \label{equ:nt_calculate}
\end{equation}
where $S_\text{IoU}$ is the IoU processing speed of the GPU, $B$ is the PCIe memory bandwidth between CPU-GPU, $\alpha$ and $\beta$ are scaling factors and $T_{\text{backbone}}$ is the processing time of the object detection backbone. Note the values of $S_\text{IoU}, B$ and $T_{\text{backbone}}$ are stable and can be obtained via system profiling. Hence, Eq. \eqref{equ:nt_calculate} provides a connection between the application-level requirements (FPS $\sim 1/T$) with the number of candidate boxes. Next, we build this hardware-adaptive relation into the learning process of robust object detectors. 

\section{Background-Attentive Adversarial Training}

Without changing the NMS module on legacy software, adversarial training (AT) is an effective way to defend against latency attacks by fundamentally screening out the phantom objects into the NMS. It solves a min-max saddle point optimization by launching attacks in the inner maximization and minimizing the AT loss in the outer optimization~\cite{advtrain}. Specifically, the goal is to learn $\theta^\ast$ under perturbations within the $l_p$-norm ball with radius $\epsilon$,
\begin{equation}
\small
\theta^\ast = \argmin_{\theta} \mathbb{E}_{(x,y)\sim \mathbb{D}} \bigl[ \max_{\norm{\delta}_p \leq \epsilon }  \mathcal{L}(f_{\theta}(x+\delta), y)\bigr]. \label{equ:min_max}
\end{equation}
However, since the existing latency attacks involve specialized loss functions (Eq. \eqref{equ:attack_loss_func}), it still remains to answer which original loss function in object detectors AT should target and what spatial region AT should attend to. 

\subsection{Objectness Loss}

Generally, object detectors learn a function $f_{\theta}(x) \rightarrow \{\mathbf{p}_k, \mathbf{b}_k\}$ to predict the probability and bounding box of $K$ objects for image $\mathbf{x} \in \mathbb{R}^{H \times W \times 3}$ with class label $\mathbf{y}$. In YOLOv5, the objective loss function can be decomposed into the classification, localization and objectness losses~\cite{yolov5}, 
\begin{eqnarray}
\small
\theta &=& \argmin_{\theta} \mathbb{E}_{\mathbf{x}\sim \mathcal{D}, \mathbf{y}, \mathbf{b}} \mathcal{L}(f_{\theta}(\mathbf{x}), \mathbf{y}, \mathbf{b}) \nonumber \\
&=& \argmin_{\theta} \mathcal{L}_{\text{cls}} + \mathcal{L}_{\text{CIoU}} + \mathcal{L}_{\text{obj}}. \label{yolov5_eq}
\end{eqnarray}
$\mathcal{L}_{\text{cls}}$ and $\mathcal{L}_{\text{CIoU}}$ emphasize on different aspects of classification and localization losses, which are exploited in~\cite{adv_misclassfy} for mis-classification and mis-location attacks. The objectness loss $\mathcal{L}_{\text{obj}}$ indicates whether a specific region contains an object in the candidate bounding box. Given the objectness confidence score $\hat{C}_i \in [0,1]$, it can be represented by the binary cross-entropy loss $\ell_{\text{BCE}}$,
\begin{equation}
\vspace{-0.03in}
\small
\mathcal{L}_{obj}= \sum_{i=1}^K\big[  \mathds{1}_i\ell_{BCE}(1,\hat{C}_i) + (1-\mathds{1}_i)\ell_{BCE}(0,\hat{C}_i)\big]. \label{equ:objloss}
\vspace{-0.03in}
\end{equation}
Recall that the adversarial loss in Eq. \eqref{equ:attack_loss_func} attempts to maximize the box confidence with auxiliary box minimization effects. Hence, we conjecture that the adversarial loss has a close relation to the objectness loss as described next.

\begin{figure}[t]
\centering
\subfloat[\footnotesize{PGD iterations}]{
\includegraphics[width=0.48\columnwidth]{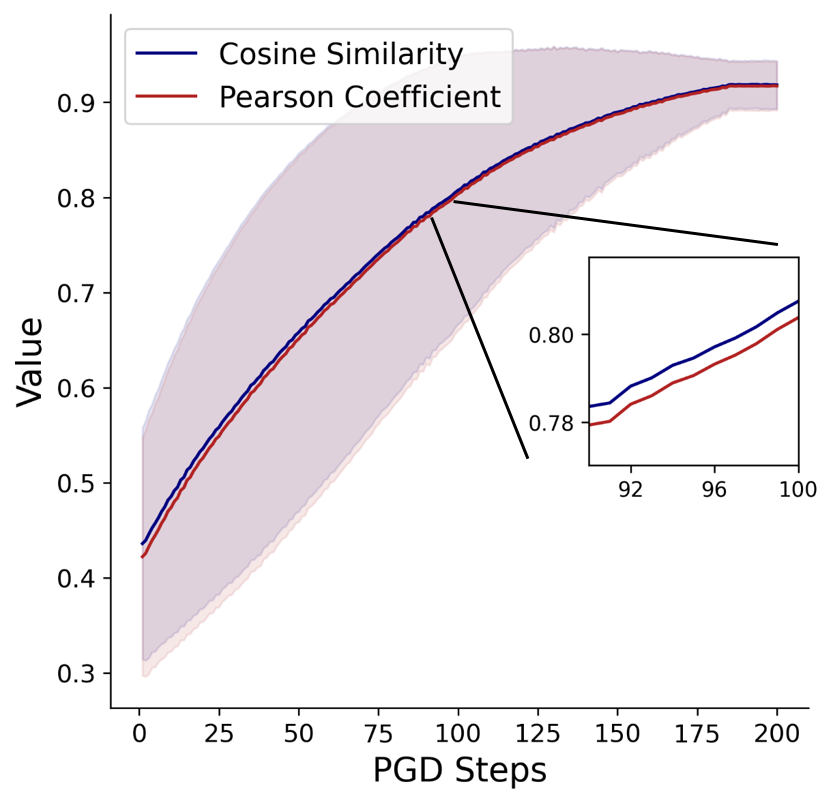}
\label{fig:cos_1}
}
\subfloat[\footnotesize{Distr. of cosine similarity}]{
\includegraphics[width=0.49\columnwidth]{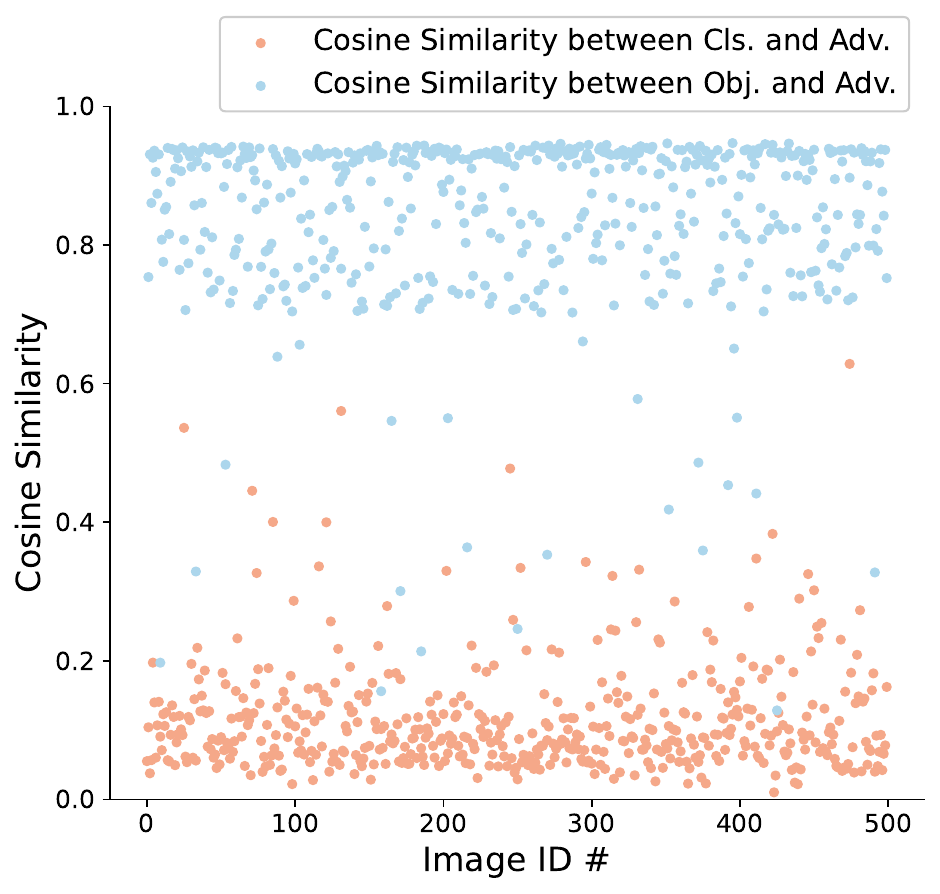}
\label{fig:cos_2}
}
\vspace{-0.05in}
\caption{\small{Adversarial relations between $\mathcal{L}_{\text{adv}}$ in~\cite{chen2024overload} and $\mathcal{L}_{\text{obj}}$. a) Pearson coefficient and cosine similarity between the two losses over the PGD iterations; b) Distribution of the cosine similarity for different images in PASCAL-VOC between $(\mathcal{L}_{\text{adv}}, \mathcal{L}_{\text{obj}})$ vs. $(\mathcal{L}_{\text{adv}}, \mathcal{L}_{\text{cls}})$.The former has stronger correlation.}}
\label{fig:cos}
\vspace{-0.2in}
\end{figure}

\textbf{Property 1 (Objectness Loss).} The adversarial perturbation $\delta'$ generated from $\mathcal{L}_{\text{adv}}$ and the perturbation $\delta$ generated from $\mathcal{L}_{\text{obj}}$ are consistent, measured by the cosine similarity
$\frac{ f_{\theta}(x+\delta') \cdot f_{\theta}(x+\delta) }{ \norm { f_{\theta}(x+\delta') }  \cdot \norm{ f_{\theta}(x+\delta) }}$, or Pearson correlation coefficient.

Fig. \ref{fig:cos} validates this empirically by tracing the evolution of averaged cosine similarity and Pearson coefficient between $f_{\theta}(x+\delta')$ and $f_{\theta}(x+\delta)$ and the distribution of the cosine similarity for different images sampled from the VOC dataset, in which $\mathcal{L}_{\text{cls}}$ is provided as contrastive examples. It is observed that as the PGD attack progresses, the two losses converge to a highly correlated adversarial subspace and the cosine similarity concentrates within a narrow band of $[0.875,0.95]$ for different images, in contrast to $\mathcal{L}_{\text{cls}}$ with weak correlation between $[0,0.2]$. This suggests that $\mathcal{L}_{obj}$ can be used as an effective \emph{proxy} in AT. Note that even if the detector does not have a separate objectness head, it could be still evidenced through the semantics to locate where objectness has been displaced, \eg, such as YOLOv8 embeds objectness inside classification~\cite{yolov8}, which is also evaluated by this work. 

An essential function of objectness is to differentiate objects from the background. Denote the objects as $x_\text{obj} = \bigcup_{k=1}^K (b_i)^{\mathbf{y}}$, $x'_\text{obj} = x_\text{obj} + \delta_\text{obj}$, and the background as $x_\text{bg} = x - x_\text{obj}, x'_\text{bg} = x_\text{bg} + \delta_\text{bg}$. We define \emph{background boundary margin} as the distance for a background region to the decision boundary of becoming a ``phantom object'' in the pixel space $\mathcal{X}$~\cite{xiao2021you},
\begin{eqnarray}
\vspace{-0.03in}
\footnotesize
&& B_\theta(x'_\text{bg}) = \min_{x'_\text{bg}} \norm{x'_\text{bg} - x}_p,  \nonumber \\
\text{s.t.} && \hat{C}_{\text{th}} - \hat{C}_i(x'_\text{bg}, b_i, y_i) = 0, \forall i \in \{1,\cdots,K \},
\vspace{-0.03in}
\end{eqnarray}
where $\hat{C}_{\text{th}}$ is the score threshold between the background and object. Similarly, $B_\theta(x'_\text{obj})$ can be derived for generating a phantom object on a natural object. 

\textbf{Property 2 (Background vs. Object Boundary Margin).} The background and object boundary margins have the following relation:
\begin{equation}
\small
\mathbb{E}_{x \sim \mathbb{D}} [B_\theta(x'_\text{bg})] \leq \mathbb{E}_{x \sim \mathbb{D}} [B_\theta(x'_\text{obj})],
\end{equation}
i.e., the perturbation strengths to generate phantoms on existing objects are larger than the background, $\norm{\delta_{\text{obj}}}_p \geq \norm{\delta_{\text{bg}}}_p$, which means that the background contains more non-robust features compared to the object regions~\cite{kim2021distilling}. 

\begin{figure}[t]
\centering
\includegraphics[width=0.98\columnwidth]{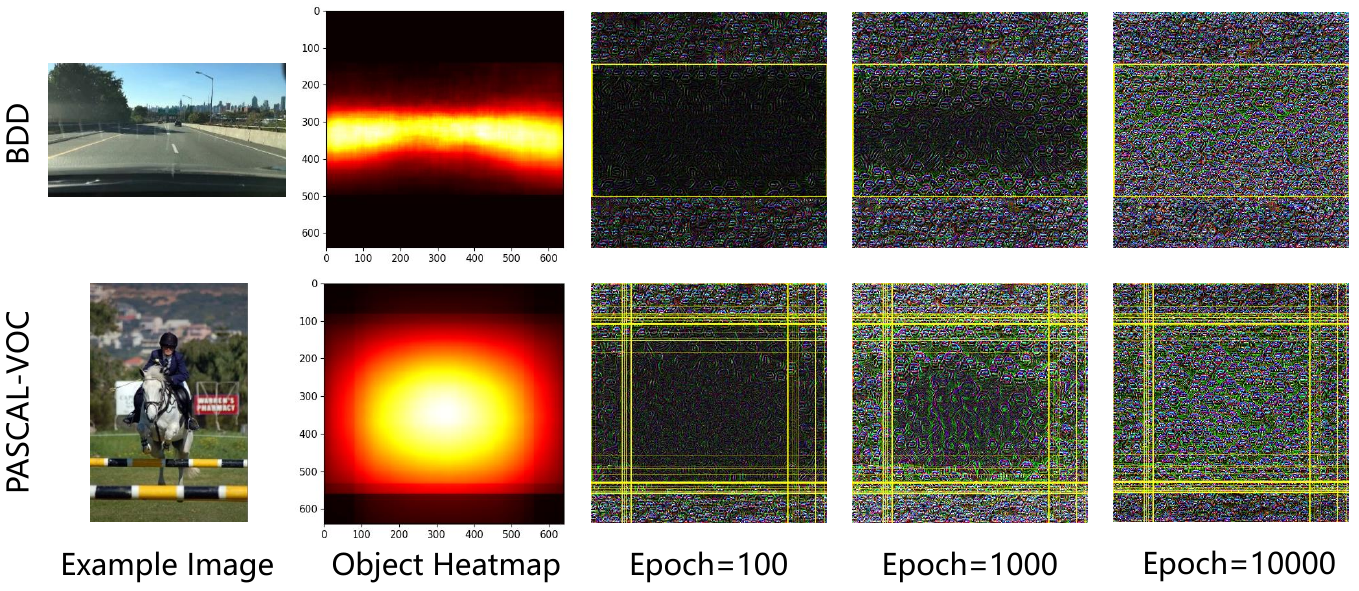}
\includegraphics[width=0.98\columnwidth]{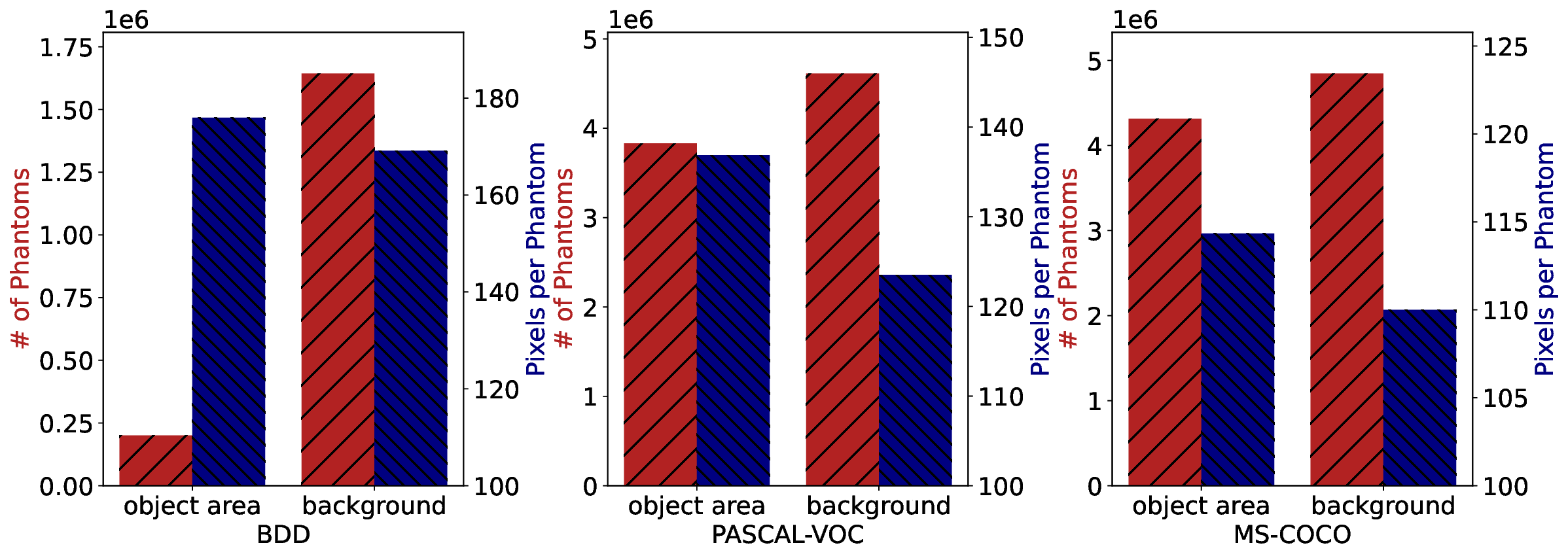}
\vspace{-0.1in}
\caption{\small{Top: Visualizing the attack processes on the object and background regions; Bottom: statistics of phantoms on different datasets. Fewer phantoms are generated on the objects compared to the background regions ({\color{red}{red bars}}). }}
\label{fig4.1:pert}
\vspace{-0.1in}
\end{figure}

Fig.~\ref{fig4.1:pert} validates this property empirically by visualizing the additive $K$-PGD process. It is observed that latency attacks initially generate perturbations in the background before the natural objects. \Eg, BDD of autonomous driving is often characterized by objects on the road so latency attacks generate objects elsewhere first. More statistics of phantoms are shown across different datasets and the background induces more phantoms with less pixel to perturb per phantom, compared to the actual objects. Since BDD usually involves small objects of vehicles and pedestrians, it is interesting to capture a sharp contrast between \# of phantoms on object vs. background. 

\begin{algorithm}[t]
\footnotesize
\caption{Background-Attentive Adversarial Training}
\label{alg:underload}
\textbf{Input}: Pre-trained model parameter $\theta_0$, dataset $\mathcal{D}$, epochs $E$, batch size $s$, attack budget $\epsilon$, learning rate $\gamma$, system metrics on R.H.S. of Eq. \eqref{equ:nt_calculate}, bounding box step size $\Delta x, \Delta y$, maximum PGD step $K$. \\
\vspace{-0.12in}
\begin{algorithmic}[1]
\STATE Initialize $|C_{\max}| \leftarrow \frac{S_\text{IoU}}{2 \alpha}\sqrt{\frac{\beta^2}{B^2}-4 \frac{\alpha}{S_\text{IoU}} (T - T_{\text{basenet}})}$,\\ $r_x, r_y \leftarrow W_0, H_0$\\
\WHILE{$ \mathbb{E}[f_\theta(\mathbf{x}+\delta^M)]_{\text{box}} < |C_{\max}|$ }
\STATE $\theta \gets \theta_0.$   \hspace{0.1in}  {\color{lightblue} $\triangleright$ re-initialize model parameter}
\FOR{epoch $\in \{1,\cdots,E\}$}   
\FOR{i $\in \{1,\cdots,K\}$} 
\STATE $\delta \gets \epsilon \cdot \text{sign}$ \big($ \nabla_\mathbf{x} \mathcal{L}_{obj}(f_{\theta}$\small($\mathbf{x}$\small)$,\mathbf{y}, \mathbf{b})$ \big) 
\STATE $\delta^M \gets \Pi_{\norm{\delta}_p \leq \epsilon} (\mathbf{M} \odot \delta)$, $\mathbf{M} = \{\mathbf{0}, \mathbf{1} \}^{r_x \times r_y}$
\STATE $\theta \gets \theta - \gamma \cdot \nabla_\theta \mathcal{L}(f_{\theta}(\mathbf{x}+\delta^M), \mathbf{y}, \mathbf{b})$
\ENDFOR
\ENDFOR
\STATE $r_x, r_y \gets r_x, r_y + \Delta x, \Delta y$ \hspace{0.1in} \hspace{0.1in}  {\color{lightblue} $\triangleright$ reduce mask size}
\ENDWHILE
\end{algorithmic}
\textbf{Output}: Obtain robust model $\theta^\ast \gets \theta$.
\end{algorithm}

\subsection{Proposed AT Method} 

Based on the unique adversarial behaviors, we develop a new AT method to improve generalization, which is orthogonal to the prior efforts in regularizing the adversarial loss surface~\cite{at1_regularize,at2_regularize}. The key insight is to make the inner optimization attend to background regions to avoid potential overfitting on object areas that necessitate much larger perturbation strengths, which also facilitates to learn more on the \emph{non-robust} regions~\cite{wang2023balance,kim2021distilling}.  

\newcommand{\graycell}{\cellcolor[rgb]{0.843,0.843,0.843}}
\begin{table*}
\small
\centering
\begin{tabular}{cccccccccccccc} 
\hline
\multirow{2}{*}{Model} & \multirow{2}{*}{Attack} & \multicolumn{3}{c}{Standard} & \multicolumn{3}{c}{MTD}  & \multicolumn{3}{c}{ODD} & \multicolumn{3}{c}{\texttt{Underload}*}  \\ 
\cmidrule(lr){3-5} \cmidrule(lr){6-8} \cmidrule(lr){9-11} \cmidrule(lr){12-14}
&   & VOC     & COCO        & BDD       & VOC    & COCO     & BDD     & VOC      & COCO    & BDD    & VOC   & COCO   & BDD   \\ \hline
\multirow{4}{*}{\rotatebox[origin=c]{90}{YOLOv3t}} & Clean          & \textcolor{red}{\textbf{55.8}} & \textcolor{red}{\textbf{36.5}} & \textcolor{red}{\textbf{30.4}} & 38.3     & 24.9     & 16.1     & 35.8     & 23.9 & 16.9     & \textcolor{blue}{48.5}         & \textcolor{blue}{25.8}         & \textcolor{blue}{18.3}          \\
      & \texttt{Daedalus}      & 2.4     & 0.2     & 0.1     & 32.3       & 17.7       & 13.2       & \textcolor{blue}{32.4}        & \textcolor{blue}{18.2}     & \textcolor{blue}{14.7}        & \textcolor{red}{\textbf{42.8}}        & \textcolor{red}{\textbf{19.3}}        & \textcolor{red}{\textbf{16.1}}         \\
      &\texttt{Phantom}        & 5.4         & 9.7         & 0.2         & \textcolor{blue}{31.7} & \textcolor{blue}{16.9} & 12.1     & 31.5     & 14.9 & \textcolor{blue}{15.3} & \textcolor{red}{\textbf{41.3}} & \textcolor{red}{\textbf{18.8}} & \textcolor{red}{\textbf{15.6}}  \\
      & \texttt{Overload}      & {\graycell} 5.2     & {\graycell}4.1     & {\graycell}0.4     & {\graycell}\textcolor{blue}{29.9}        & {\graycell}\textcolor{blue}{16.5}        & {\graycell}12.0       & {\graycell}29.4       & {\graycell}16.4        & {\graycell}\textcolor{blue}{16.3}        & {\graycell}\textcolor{red}{\textbf{33.3}}        & {\graycell}\textcolor{red}{\textbf{18.7}}        & {\graycell}\textcolor{red}{\textbf{16.8}}         \\ 
\hline
\multirow{4}{*}{\rotatebox[origin=c]{90}{YOLOv5s}} & Clean         & \textcolor{red}{\textbf{73.3}} & \textcolor{red}{\textbf{51.3}} & \textcolor{red}{\textbf{50.7}} & 57.7     & 40.5     & 34.0     & 55.9     & 38.1 & 33.8     & \textcolor{blue}{68.9}         & \textcolor{blue}{43.6}         & \textcolor{blue}{34.6}          \\
      & \texttt{Daedalus}      & 12.8    & 15.1    & 6.4     & \textcolor{blue}{50.1}        & \textcolor{blue}{33.0}        & \textcolor{blue}{28.1}        & 48.4       & 32.8        & 27.6       & \textcolor{red}{\textbf{50.3}}        & \textcolor{red}{\textbf{33.4}}        & \textcolor{red}{\textbf{28.8}}         \\
      &\texttt{Phantom}        & 7.5         & 8.8         & 7.6         & \textcolor{blue}{54.7} & \textcolor{blue}{36.8} & \textcolor{blue}{24.7} & 53.3     & 36.5 & 24.5     & \textcolor{red}{\textbf{61.3}} & \textcolor{red}{\textbf{36.9}} & \textcolor{red}{\textbf{25.1}}  \\
      & \texttt{Overload}      & {\graycell}4.5     & {\graycell}7.9     & {\graycell}4.7     & {\graycell}\textcolor{blue}{49.4}        & {\graycell}36.5       & {\graycell}24.5       & {\graycell}49.2       & {\graycell}\textcolor{blue}{35.8}     & {\graycell}\textcolor{blue}{25.8}        & {\graycell}\textcolor{red}{\textbf{53.3}}        & {\graycell}\textcolor{red}{\textbf{36.7}}        & {\graycell}\textcolor{red}{\textbf{26.5}}         \\ 
\hline
\multirow{4}{*}{\rotatebox[origin=c]{90}{YOLOv8s}} & Clean          & \textcolor{red}{\textbf{82.5}} & \textcolor{red}{\textbf{57.7}} & \textcolor{red}{\textbf{53.4}} & 68.8     & 43.6     & 36.1     & 68.7     & 42.2 & 37.1     & \textcolor{blue}{72.6}         & \textcolor{blue}{45.7}         & \textcolor{blue}{39.3}          \\
      & \texttt{Daedalus}      & 18.8    & 16.1    & 2.6     & \textcolor{blue}{65.1}        & \textcolor{blue}{41.8}        & \textcolor{blue}{30.1}        & 64.3       & 40.7        & 26.9       & \textcolor{red}{\textbf{69.7}}        & \textcolor{red}{\textbf{43.6}}        & \textcolor{red}{\textbf{33.2}}         \\
      &\texttt{Phantom}        & 6.8         & 10.9        & 3.5         & 61.3     & \textcolor{blue}{41.9} & \textcolor{blue}{34.5} & \textcolor{blue}{64.6} & 40.3 & 34.4     & \textcolor{red}{\textbf{66.1}} & \textcolor{red}{\textbf{42.3}} & \textcolor{red}{\textbf{36.5}}  \\
      & \texttt{Overload}      & {\graycell}7.7     & {\graycell}20.1    & {\graycell}5.7      & {\graycell}53.0       & {\graycell}\textcolor{blue}{40.6}        & {\graycell}34.3       & {\graycell}\textcolor{blue}{62.4}        & {\graycell}39.1        & {\graycell}\textcolor{blue}{34.9}        & {\graycell}\textcolor{red}{\textbf{66.5}}        & {\graycell}\textcolor{red}{\textbf{42.0}}        & {\graycell}\textcolor{red}{\textbf{36.0}}         \\
\hline
\end{tabular}
\caption{\small{Defense performance (mAP50 \%) under the latency attacks of \texttt{Daedalus}~\cite{wang2021daedalus}, \texttt{Phantom Sponge}~\cite{ps} and \texttt{Overload}~\cite{chen2024overload} while comparing with existing defense of MTD and OOD. The {\textbf{\color{red} Best}} and {\color{blue} Second Best} values in each row are marked in {\textbf{\color{red} Red}} and {\color{blue} Blue}. The first rows of ``Clean'' compare the clean accuracy drop with different AT methods.}}
\label{tab:main_results}
\end{table*}

We adopt a binary mask $\mathbf{M} \in \{\mathbf{0}, \mathbf{1} \}^{r_x^i \times r_y^i}$ to control the amount of perturbation being injected into the inner maximization. $r_x^i, r_y^i$ are the width and height of an object $i$. Our goal is to learn a robust model $\theta^\ast$ by minimizing the overall loss function $\mathcal{L}$ defined by Eq. \eqref{yolov5_eq} such that the latency attacks are suppressed under the hardware capacity. The optimization is formalized as,
\begin{equation}
\small
\theta^\ast = \argmin\limits_{\theta, {\delta^M \in \mathcal{S}} }\mathcal{L} \big(f_{\theta}(\mathbf{x}+\delta^M), \mathbf{b}, \mathbf{y} \big) \label{obj}
\vspace{-0.05in}
\end{equation}
where

\begin{equation}
\small
\mathcal{S} = \bigl\{\delta^M \leftarrow ~\mathbf{M} \odot \argmax\limits_{\norm{\delta}_p \leq \epsilon, i\in \mathcal{K}} \mathcal{L}_{obj}(f_{\theta}(\mathbf{x}+\delta), b_i, y_i) \bigr\} \label{constraint1}
\end{equation}
\begin{equation}
\small
\mathbb{E}[f_\theta(x+\delta^M)]_{\text{box}} < \frac{S_\text{IoU}}{2 \alpha}\sqrt{\frac{\beta^2}{B^2}-4 \frac{\alpha}{S_\text{IoU}} (T - T_{\text{backbone}})}    \label{constraint2} \\
\end{equation}
\begin{equation}
\small
\mathbf{M} = \{\mathbf{0}, \mathbf{1} \}^{r_x^i \times r_y^i}, r_x^i \in [0,W_b^i], r_y^i \in [0,H_b^i], i\in\mathcal{K}. \label{constraint3}
\end{equation}
Eq. \eqref{constraint1} finds the set of background attentive adversarial perturbations $\mathcal{S}$ targeting the objectness loss, in which $\odot$ is the Hadamard product. Eq. \eqref{constraint2} states that the number of candidate boxes in the object detector generated by masked perturbation $\delta^M$ should be less than the hardware capacity obtained from Eq. \eqref{equ:nt_calculate}. This successfully connects the optimization process with the hardware capacity on edge systems with a stopping criteria. Eq. \eqref{constraint3} bounds the $0$-$1$ mask generation with the width $W_b^i$ and height $H_b^i$ of an object. 

\textbf{Property 3 (Monotonicity).} The box count $\mathbb{E}[f_\theta(x+\delta^M)]_{\text{box}}$ is monotonously increasing regarding the mask size $r_x^i \times r_y^i$. 

Based on this property, we develop an efficient algorithm to learn a robust model $\theta^\ast$ as long as the box count is below hardware capacity. The details are shown in Algorithm \ref{alg:underload}. Starting with a pre-trained model, we initialize the mask sizes to the maximum size of $W_b \times H_b$. Then we perform $E$ epochs of AT before we reduce the mask with step sizes of $(\Delta x, \Delta y)$. The iteration stops until we reach the hardware capacity. More experiments of the robust vs. clean accuracy are available in the next section.

\section{Experiments} \label{sec:evaluation}
\subsection{Experimental Settings}
\noindent\textbf{Datasets \& Models.} 
We conduct experiments on three standard datasets including PASCAL-VOC~\cite{voc}, MS-COCO~\cite{coco}, and Berkeley DeepDrive (BDD)~\cite{bdd100k}. 
PASCAL-VOC and MS-COCO are common benchmarks in object detection, while BDD is frequently used for autonomous driving. 
We follow the standard ``07+12'' protocol on VOC. For MS-COCO, we train on the \texttt{train+valminusminival} 2017 set and test on the \texttt{minival 2017 test}. For BDD, we use BDD100K with 70K training and 10K testing samples. 

We evaluate three models in our evaluation including YOLOv3~\cite{yolov3}, YOLOv5~\cite{yolov5} and YOLOv8~\cite{yolov8}. The existing attacks primarily target anchor-based models such as YOLOv3 and YOLOv5~\cite{ps,chen2024overload}. Our work extends this to cover the latest anchor-free YOLOv8 models~\cite{yolov8}, which provides a holistic coverage of mainstream object detectors for edge devices across different YOLO generations. 

\noindent\textbf{Attack and Defense Settings.} 
We evaluate our defense against three existing attacks: \texttt{Daedalus}~\cite{wang2021daedalus}, \texttt{Phantom Sponge}~\cite{ps} and \texttt{Overload}~\cite{chen2024overload}. We set the hyperparameters following~\cite{ps}, $\lambda_1=1, \lambda_3=0$ with the $l_2$ norm and utilize $K$-step PGD ($K=4$) with perturbation magnitude of $\epsilon=1/255$ for AT~\cite{advtrain}. Starting from the pre-trained model, we employ AdamW~\cite{adamw} as the optimizer with scheduled learning rate and follow~\cite{hyp} to set other hyperparameters for reproducibility. 

We compare the proposed \texttt{Underload} with the SOTA defense in object detection: the multi-task domain defense (MTD)~\cite{mtd} and objectness-oriented defense (OOD)~\cite{objadv}. MTD leverages the asymmetric role of task losses for improving robustness with AT. OOD develops an attack against the objectness loss and performs AT in an identical manner to MTD.

\subsection{Robustness Evaluation}
The primary results are available in Table~\ref{tab:main_results} measured by mAP50, a metric representing the average precision across all classes at the 50\% IoU threshold. For each model, the rows represent the model accuracy under ``Clean'' (no attack) followed by three latency attacks. In each column, we compare the defense performance of Standard (no defense), MTD and OOD with \texttt{Underload}. Although MTD is not specifically tailored against latency attacks, it retains certain defensive capabilities due to the intertwined nature of loss functions in multi-tasking object detectors. 

By tracing through the values marked in red color, we can see that \texttt{Underload} demonstrates superior performance in both clean and robust accuracy compared to existing defense. \Eg, on YOLOv5s, \texttt{Underload} brings the robust accuracy from 7.5\% back to 61.3\% under the \texttt{Phantom Sponge} attack and from 4.5\% back to 53.3\% under the \texttt{Overload} attack for VOC, with only 4.4\% drop of clean accuracy, compared to 15.6\% and 17.4\% drop using MTD and OOD. This is because \texttt{Underload} has taken active measures to only inject useful perturbations into the AT process. MTD could screen out some attacks but its performance is not stable on the VOC-YOLOv3t and VOC-YOLOv8s pairs.  
OOD does not consider the balance between clean accuracy and robust accuracy, thus suffering from more clean accuracy drops compared to \texttt{Underload}. BDD dataset features small objects such as vehicles and pedestrians on the road that leave a large background space open for the attack -- \texttt{Underload} also achieves 1-3\% higher clean/robust accuracy under these hard-to-defense cases than the benchmarks.

\begin{figure}[t]
\centering
\subfloat[\footnotesize{Jetson Xavier NX}]{
\includegraphics[width=0.49\columnwidth]{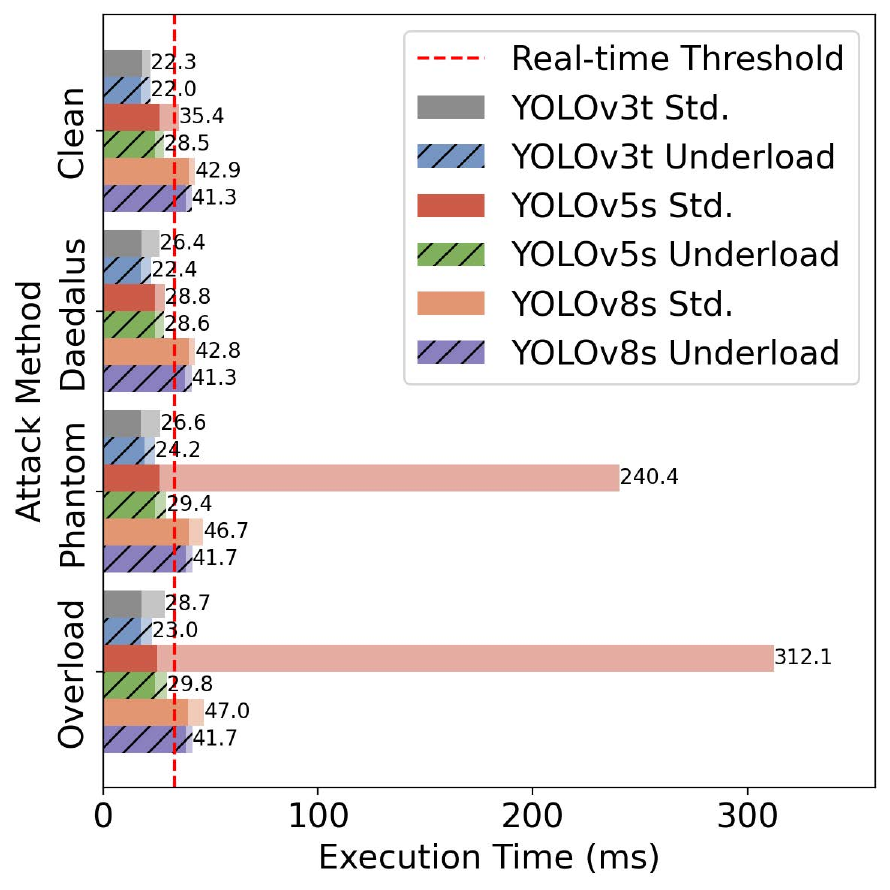}
\label{fig:xavier}
}
\subfloat[\footnotesize{Jetson Orin NX}]{
\includegraphics[width=0.49\columnwidth]{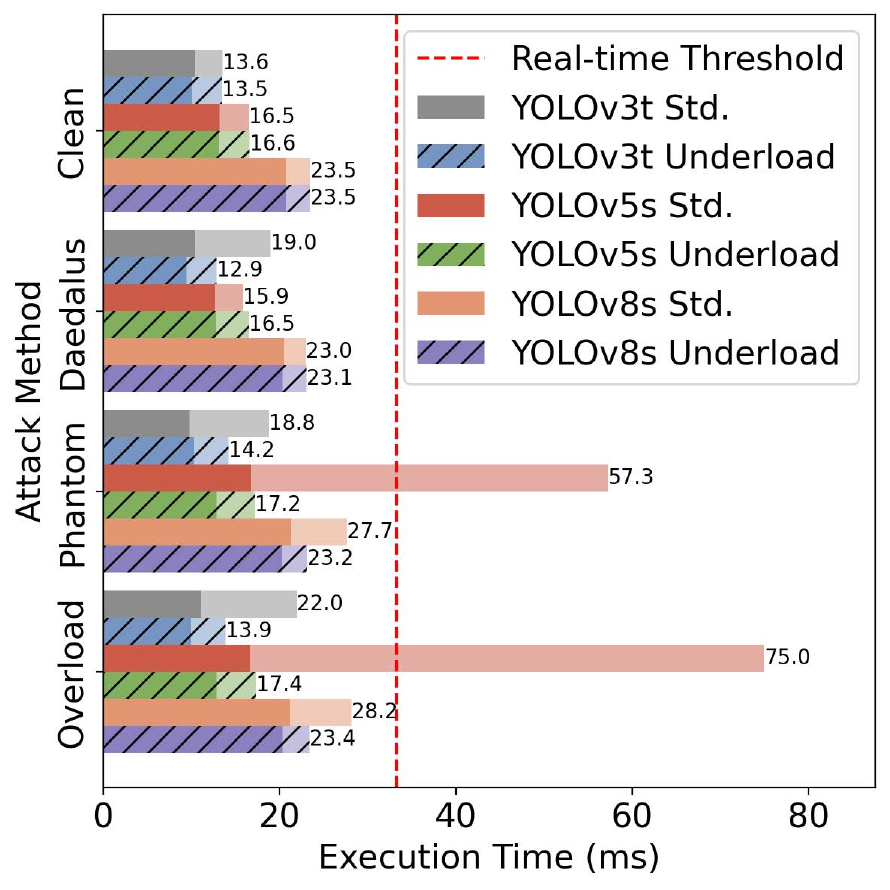}
\label{fig:orin}
}
\\
\subfloat[\footnotesize{4070 Ti Super}]{
\includegraphics[width=0.49\columnwidth]{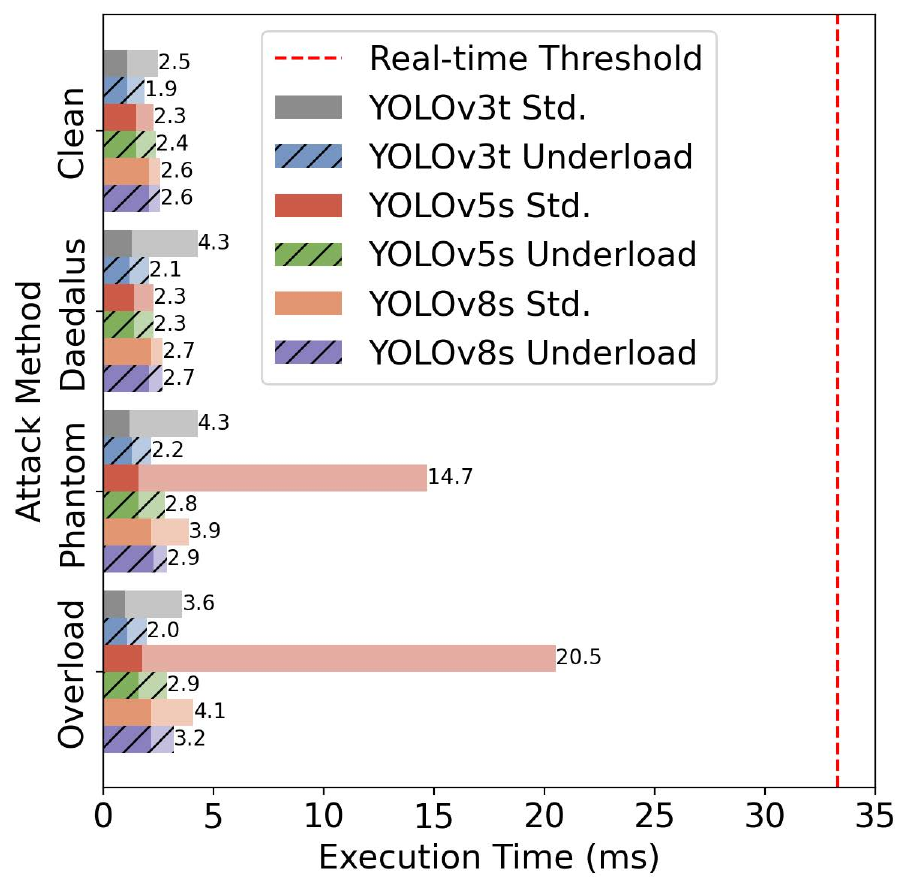}
\label{fig:4070}
}
\subfloat[\footnotesize{Nvidia A100}]{
\includegraphics[width=0.49\columnwidth]{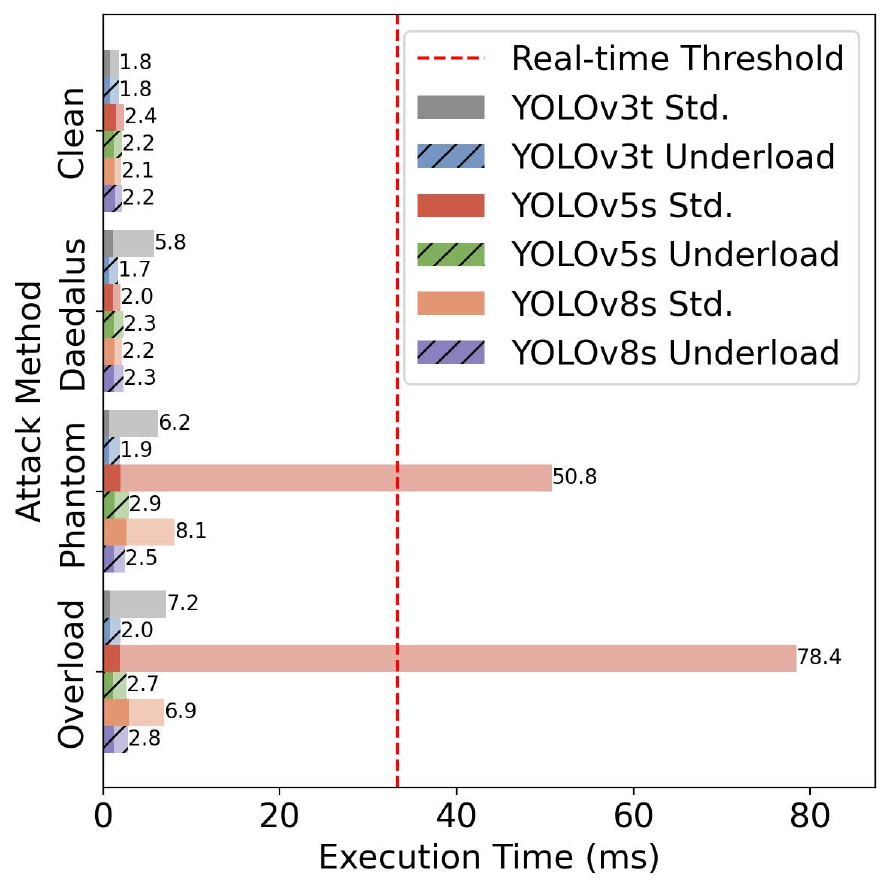}
\label{fig:a100}
}
\vspace{-0.05in}
\caption{\small{Execution time (ms) across heterogeneous GPUs under different attacks and effectiveness of our defense.}}
\vspace{-0.2in}
\label{fig:defense_restore}
\end{figure}

\subsection{Evaluation on Heterogeneous GPUs}
Next, we demonstrate how our defense restores the real-time processing capabilities across heterogeneous devices of edge, desktop and server-grade GPUs. Fig. \ref{fig:defense_restore} shows the total execution time of the basenet and NMS on VOC. First, we can see that \texttt{Phantom Sponge} and \texttt{Overload} are the two strongest attacks that successfully push processing time over the real-time threshold (30 FPS), even on A100 GPUs. This validates that the true bottleneck migrates from computation to memory on high-end GPUs, which is still a serious problem though the computational power is sufficiently high. Further, when memory bandwidth is shared on multi-tenant cloud platforms, latency attacks become more effective on A100 by comparing with the 4070Ti's single-tenant setup. Fortunately, our defense restores the processing time of all the cases close to the original states (clean). \Eg, on Jetson Orin NX, we successfully restore the latency back to 13, 17, and 23 ms for different YOLO generations that meet the 30 FPS requirements for most end-user applications.

\subsection{Ablation Study of Mask Size}

\begin{figure}[t]
\centering
\subfloat[\footnotesize{Box \# vs. Mask Ratio}]{
\includegraphics[width=0.48\columnwidth]{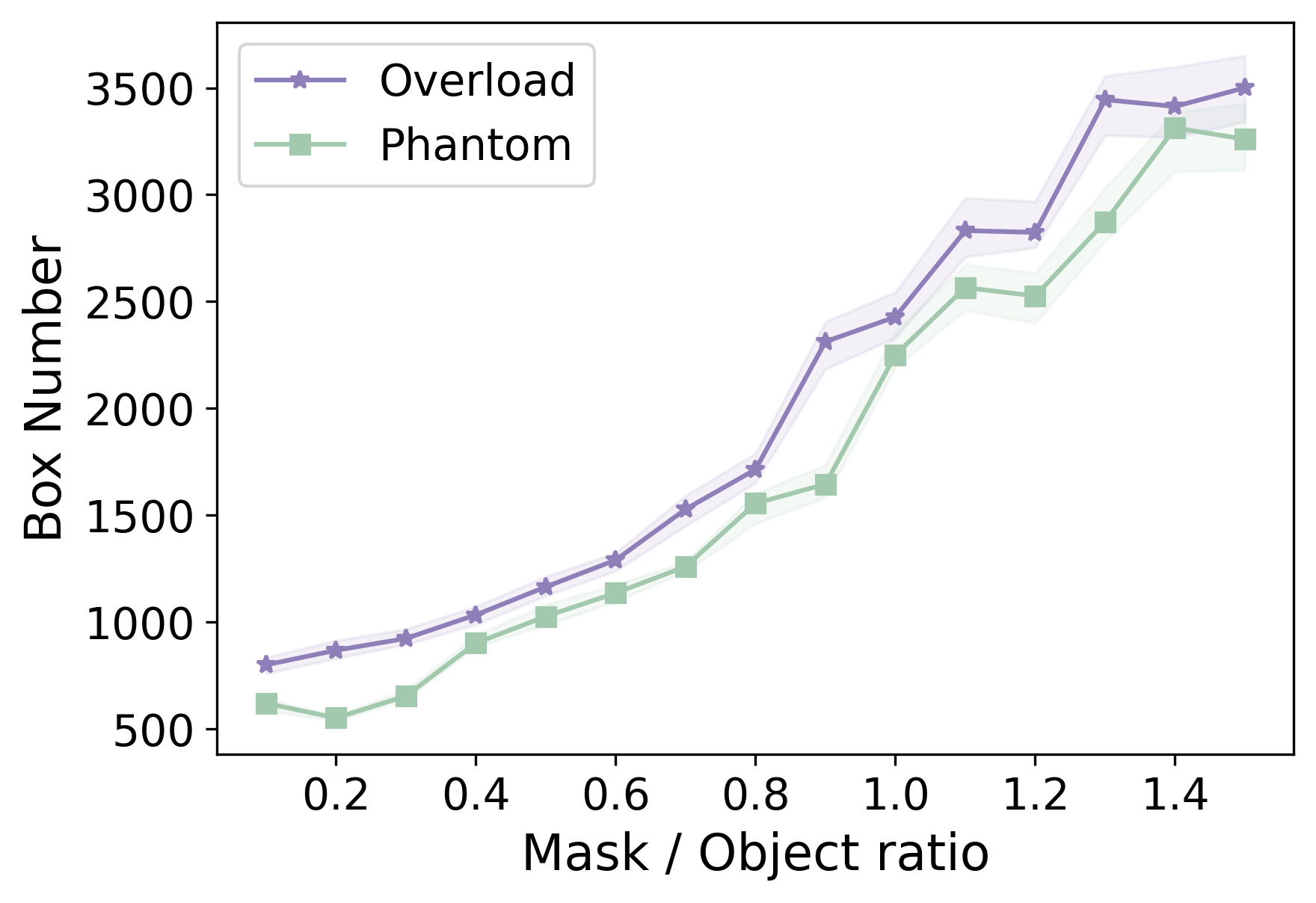}
\label{fig:ab1}
}
\subfloat[\footnotesize{Clean vs. Robust Acc}]{
\includegraphics[width=0.49\columnwidth]{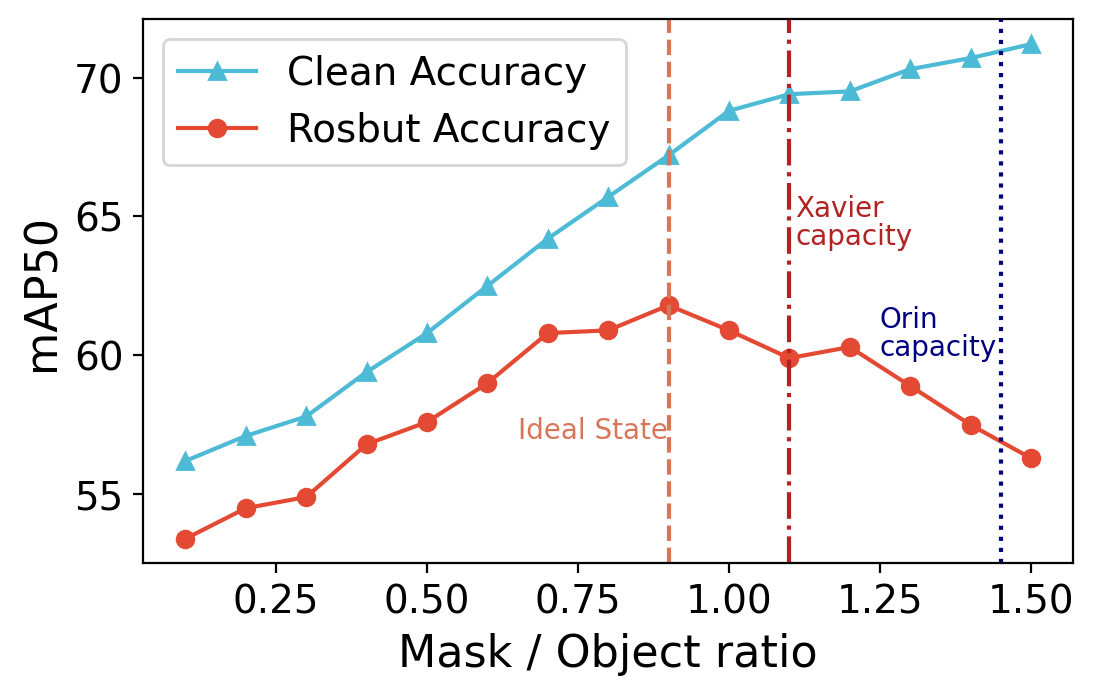}
\label{fig:ab2}
}
\\
\subfloat[\footnotesize{Perturbation generated by Overload for different ratios.}]{
\includegraphics[width=0.96\columnwidth]{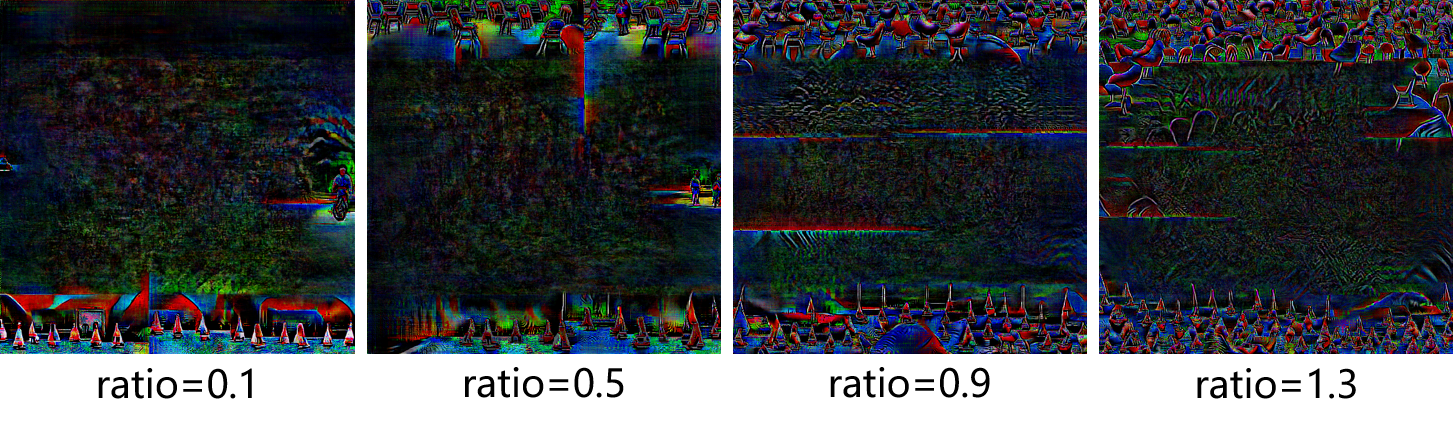}
\label{fig:pert_vis}
}
\vspace{-0.1in}
\caption{\small{Ablation study of mask size for clean accuracy and robustness on PASCAL-VOC.}}
\vspace{-0.2in}
\label{fig:ablation_study}
\end{figure}

Mask size plays a pivotal role to steer the attention of AT, thus potentially balancing the clean and robust accuracy. Fig.\ref{fig:ablation_study} illustrates this relation by examining the mask/object ratio, which varies from $10$\% to $150$\% on the x-axis. 
First, it is obvious from Fig.\ref{fig:ab1} that the box number increases with the mask size (unprotected area). In Fig.\ref{fig:ab2}, the clean accuracy climbs continuously because larger mask sizes reduce the amount of perturbations injected into the inner optimization, that helps improve generalization reflected on the clean accuracy. Robust accuracy exhibits a more interesting pattern, which increases with the mask/object ratio and peaks around $0.9$ for VOC. This is counter-intuitive as the initial segments of the robust curve should have decreased when the protected area shrinks (increment of mask size). 

We conjecture that if a mask is too small compared to natural objects, it is difficult to generate phantom objects in the mask area. Hence, the mask size could be raised close to the contour of natural objects for maximizing the attack capacity but not sacrificing the clean accuracy too much. Fig.\ref{fig:pert_vis} visualizes the phantom perturbation generated from the mask ratio of $\{0.1,\cdots,1.3\}$. We can see that phantoms of repetitive patterns are being generated in the letterbox area first. Background regions with a larger mask ratio are filled with more phantoms because the unprotected area is also larger. 

\begin{figure}
\centering
\subfloat[Original image]{\includegraphics[width=0.32\columnwidth]{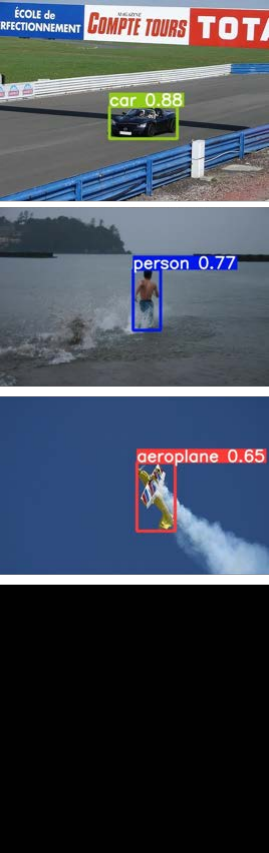}\label{fig_s:1.1.a}}
\hspace{0.5pt}
\subfloat[Targeting ``car'']{\includegraphics[width=0.32\columnwidth]{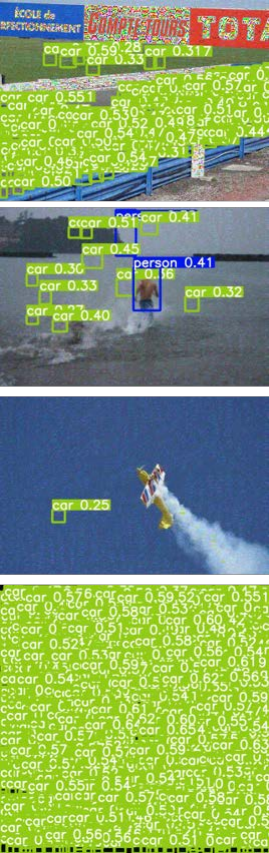}\label{fig_s:1.1.b}} 
\hspace{0.5pt}
\subfloat[Targeting ``boat'']{\includegraphics[width=0.32\columnwidth]{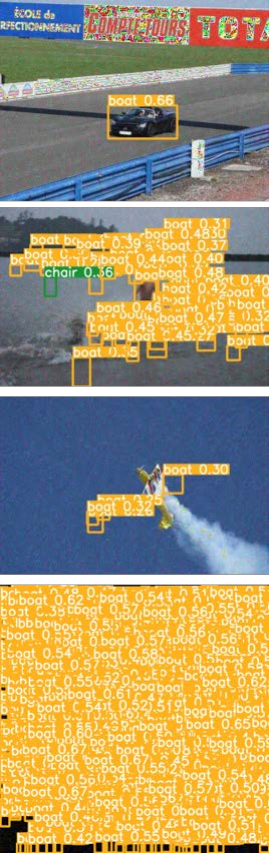}\label{fig_s:1.1.c}} 
\caption{\small{Validating the background and object semantics by \emph{targeted attacks} of creating phantom ``cars'' and ``boats'': ``\#1: a racing car on the track'', ``\#2: a person on the beach'', ``\#3: an aeroplane in the sky'', and ``\#4: a black image with no information''. More ``cars'' are generated than ``boats'' on the road and opposite is found on the beach. The blank image is used for comparison as it contains no semantic information.}}
\label{fig_s_1:seg}
\vspace{-0.15in}
\end{figure}

\subsection{Background and Object Semantics}

Finally, we expose intriguing artifacts between the background and object semantics through the lens of \emph{targeted attacks}, that echo with our design of background-attentive AT. The goal is to launch targeted attacks to generate phantom objects of ``cars'' and ``boats'' under different background regions to different images in Fig.~\ref{fig_s_1:seg}. For \#1, it is much easier to generate more ``cars'' on the track than ``boats'' and the opposite is true for \#2. For \#3, it is a bit easier to generate more ``boats'' in the sky than ``cars'', due to the blue color of the sky similar to the ocean. For \#4, it is equally possible to generate as many ``cars'' or ``boats'' since the blank image contains no semantic information.

From the attacker's perspective, the attack success rate can be improved by targeting the closest semantics to the background region, \Eg, generating phantom ``birds'' in the ``sky'', or ``persons'' on the ``road'', which not only requires less perturbation (stealthier), but also generates more phantoms with higher latency in the NMS. This is because the semantically close objects/background are more vulnerable to adversarial attacks that slightly push the objects/background region over the manifold to be adversarial. From the defense perspective, with prior knowledge about the application domain (training set), it is also possible to finetune AT against specific semantic classes (vulnerable classes).

\section{Conclusion}
This work restores the real-time processing capabilities back to object detectors under latency attacks. We leverage background-attentive AT to focus more on the vulnerable background regions, and bring hardware capacity into the robust learning process. The extensive experiments demonstrate effectiveness to defend against latency attacks on various datasets, models and GPUs. 

\small{\textbf{Acknowledgment.} This work is supported in part by NSF of Zhejiang Province LZ25F020007, NSFC 62394341, 92467301, 62293511, the Fundamental Research Funds for the Central Universities
226202400182 and the ZJUCSE-Enflame Cloud and Edge Intelligence Joint Laboratory.}

{
    \small
    \bibliographystyle{ieeenat_fullname}
    \bibliography{main}
}

\clearpage

\setcounter{page}{1}
\maketitlesupplementary
\renewcommand{\thesection}{\Alph{section}}
\setcounter{section}{0}

\begin{figure*}[ht]
\centering
\includegraphics[width=0.9\textwidth]{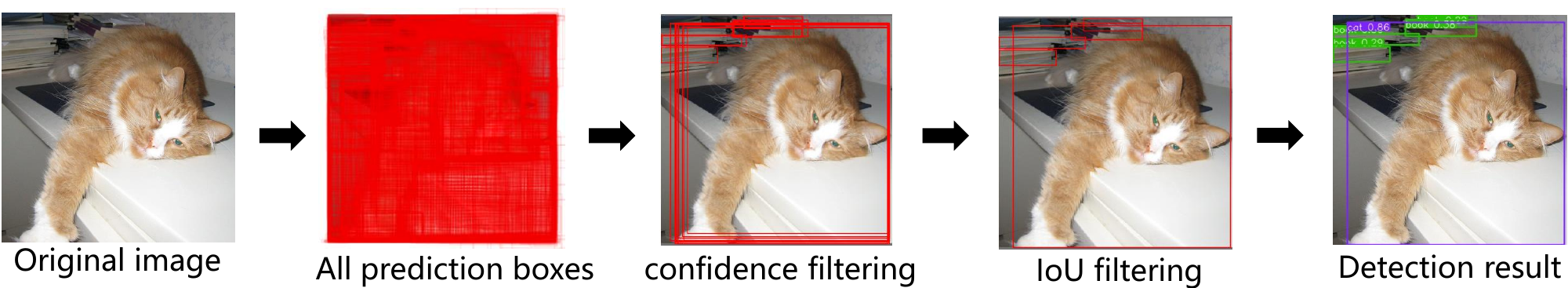}
\caption{Visualization of the entire NMS process: a) original image; 2) pre-processed results with all the prediction boxes; 3) box filtering with confidence threshold; 4) additional filtering with IoU threshold; 5) Final detection result.}
\label{fig:nms}
\end{figure*}

\section{More examples of NMS and Latency Attacks}
\label{sec:nms}

During the training phase, object detectors such as YOLO~\cite{yolov3,yolov5} and Faster-RCNN~\cite{fastercnn} usually apply a many-to-one matching strategy, that the prediction results contain multiple detection boxes for the same object with redundancy. The NMS module removes this redundancy by reducing the number of detection boxes to balance the precision and recall. 
As shown in Fig.~\ref{fig:nms}, the model predicts a number of boxes to detect the object of cat. The box number is related to the hyperparameters such as the number of anchors. The initial confidence filtering removes the most irrelevant background bounding boxes.

The primary goal of latency attacks is to ensure that the confidence scores of most bounding boxes exceed the confidence threshold. Unlike another type of attack that solely targets model efficiency~\cite{ilfo, nmtsloth}, latency attacks on object detectors not only reduces the processing speed, but also the detection accuracy. Therefore, its defense carries greater practical value. 

\begin{figure}[t]
\centering
\subfloat[DETR inference on 4070ts]{\includegraphics[width=0.48\columnwidth]{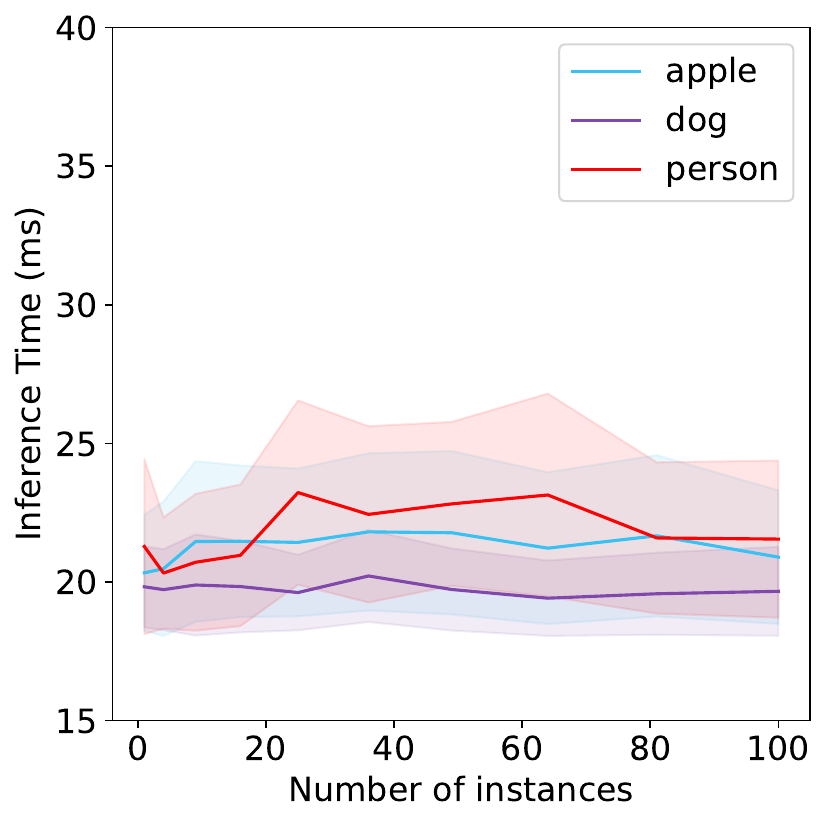}\label{fig_s:detr_4070ts}}
\hspace{0.5pt}
\subfloat[RT-DETR inference on 4070ts]{\includegraphics[width=0.48\columnwidth]{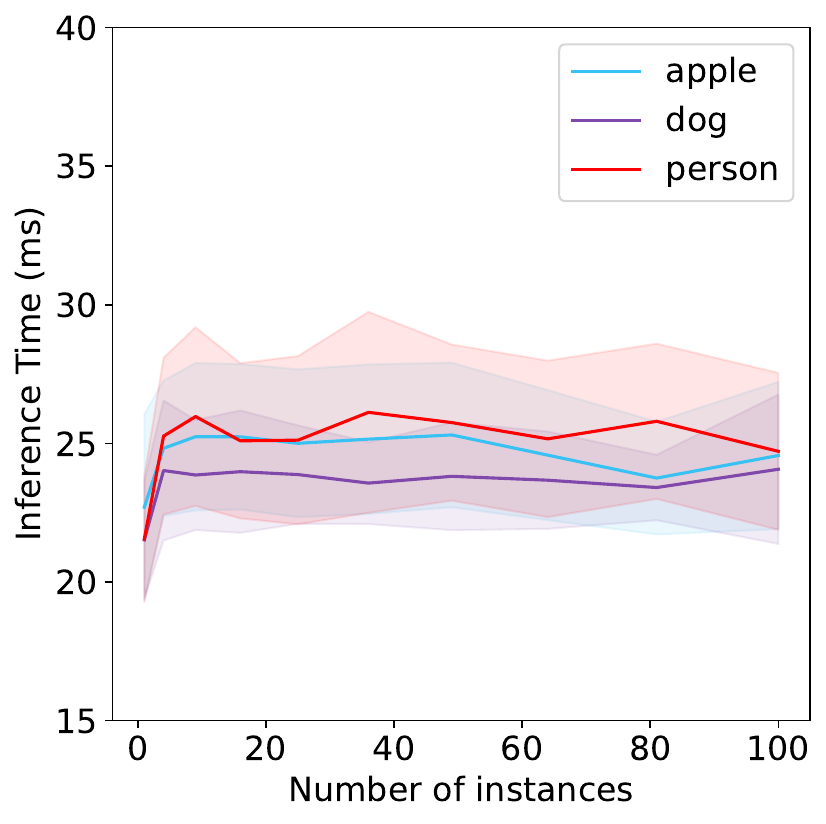}\label{fig_s:rtdetr_4070ts}} 
\hspace{0.5pt}
\subfloat[DETR inference on Orin NX]{\includegraphics[width=0.48\columnwidth]{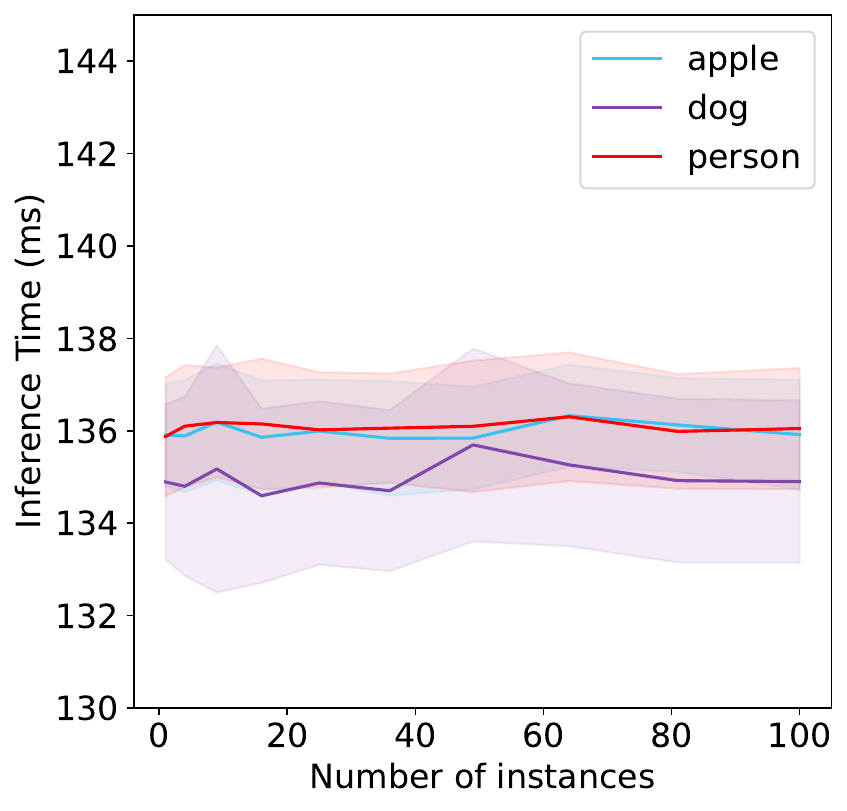}\label{fig_s:detr_orin}}
\hspace{0.5pt}
\subfloat[RT-DETR inference on Orin NX]{\includegraphics[width=0.48\columnwidth]{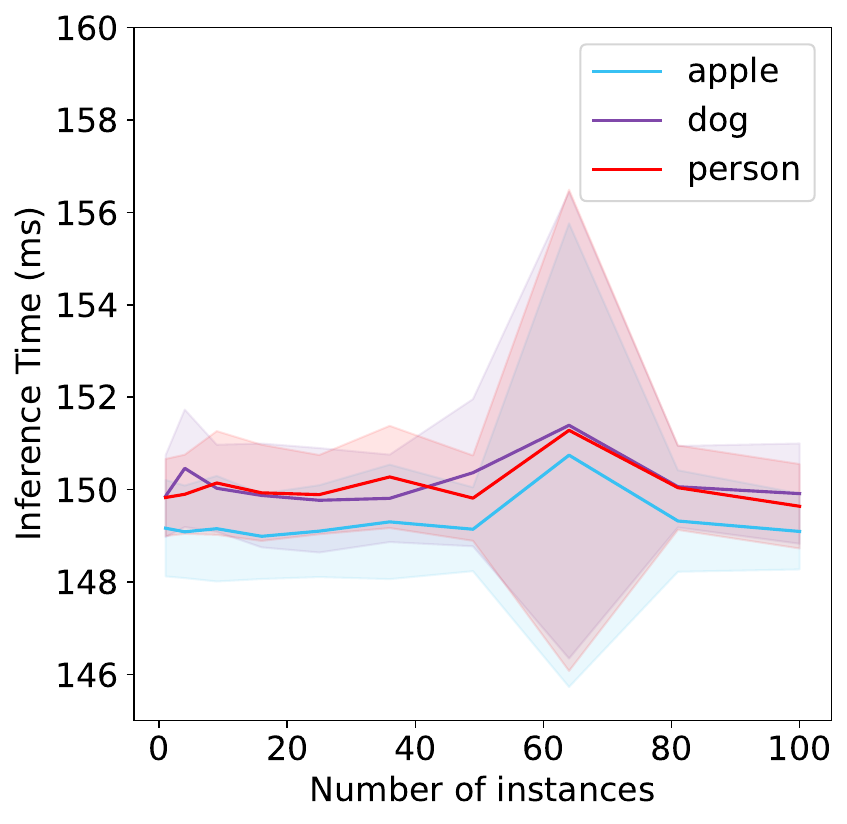}\label{fig_s:rtdetr_orin}} 
\hspace{0.5pt}
\caption{\small{DETR and RT-DETR inference time evaluation on different devices.}}
\label{fig_s:detr_inference}
\end{figure}

\section{When DETR meets Latency Attacks} \label{sec:detr}

DETR (Detection Transformer) utilizes the Hungarian algorithm for one-to-one matching between predicted boxes and ground truth boxes, enforcing a strict limit on the number of detection boxes (e.g., 100 boxes)~\cite{dert}. Intuitively, DETR should be agnostic to the number of objects. In terms of robustness to perturbations, the previous works also suggest that transformers such as ViT exhibit much higher robustness against gradient-based PGD attacks compared to CNN models~\cite{understanding}. 

These have collectively led to the following question: \emph{whether the DETR families have the similar latency attack surface as the CNN-based object detectors?} To answer this question, we analyze the performance variations with the number of instances ranging from $[0,100]$. We also investigate the latest advance from RT-DETR (Real-Time Detection Transformer)~\cite{detr_beats_yolo} under the pressure of latency attacks.

We select images with a single instance from three categories as the candidates, placing each image into an $N \times N$ grid to generate images with $N^2$ instances. We employ DETR and RT-DETR to perform inference on these images 100 times and examine the average execution time on Nvidia 4070Ti Super and Jetson Orin NX in Fig~\ref{fig_s:detr_inference}. 

The preliminary experimental results show that execution time does not vary significantly with different number of instances. From an architectural design perspective, the stability is because DETR predicts all objects from end-to-end without an additional hand-craft NMS module for redundant box elimination. Therefore, it is tempting to conclude that, as long as the matching threshold/number of boxes has been set under the hardware capacity, DETRs do not expose the same vulnerability to latency attacks as their CNN counterparts.

\section{More Details of Experimental Settings}
We perform all the AT on a workstation equipped with 8 Nvidia RTX 4090 GPUs, Intel Xeon Gold 6326 CPUs and 480 GB of RAM. The Python environment used for training and validation is configured with Python 3.9, PyTorch 2.0.0, Torchvision 0.15.0, and CUDA version 11.7. The edge device utilized Python 3.8, PyTorch 2.0.0, Torchvision 0.15.0, and CUDA version 11.4. The AT implementation for YOLOv3 and YOLOv5 uses the Ultralytics-YOLOv5 interface~\cite{yolov5}, while YOLOv8 adopts the Ultralytics interface~\cite{yolov8}. Hyperparameter selection are described in the main text of the paper, and the experiments are repeated for $5$ times.

\begin{figure*}[t]
\centering
\includegraphics[width=0.95\textwidth]{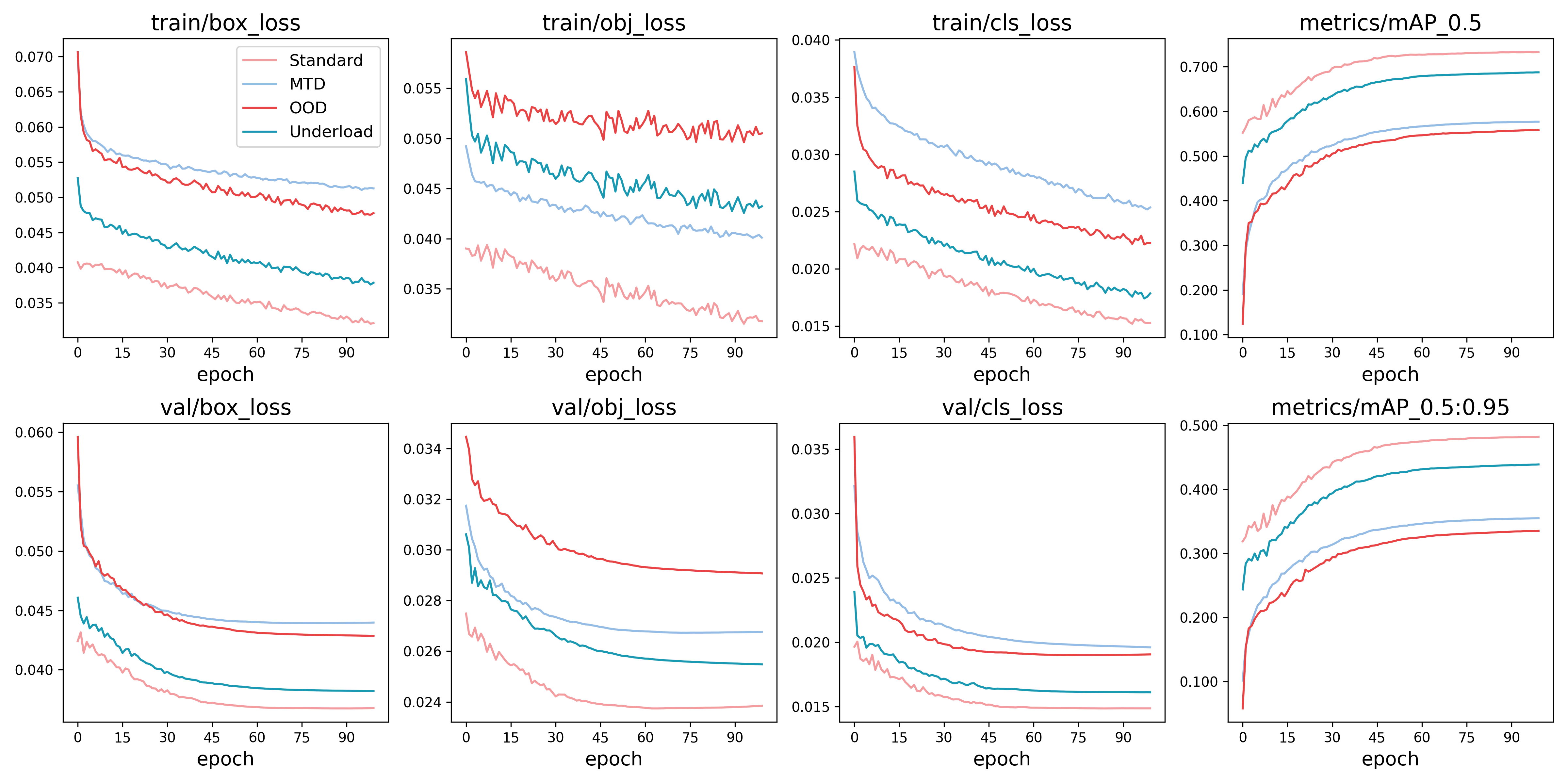}
\caption{\small{Visualization of the training process of YOLOv5s on the PASCAL-VOC dataset.}}
\label{fig_s_1:training}
\vspace{-0.15in}
\end{figure*}

\section{Analysis of Training Process}
\begin{figure*}[t]
\centering
\subfloat[Standard]{\includegraphics[width=0.45\textwidth]{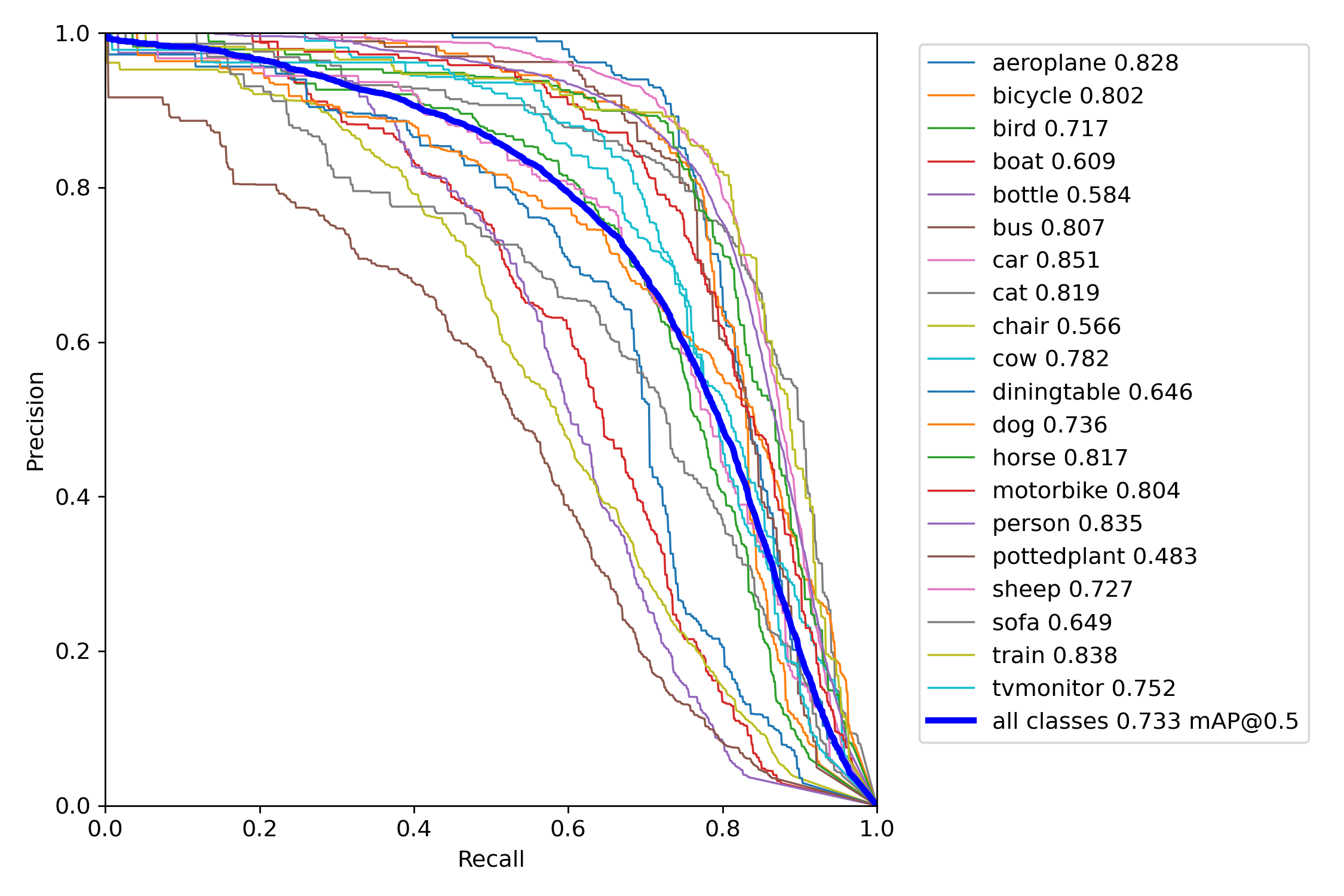}\label{fig_s:2.a}}
\subfloat[MTD]{\includegraphics[width=0.45\textwidth]{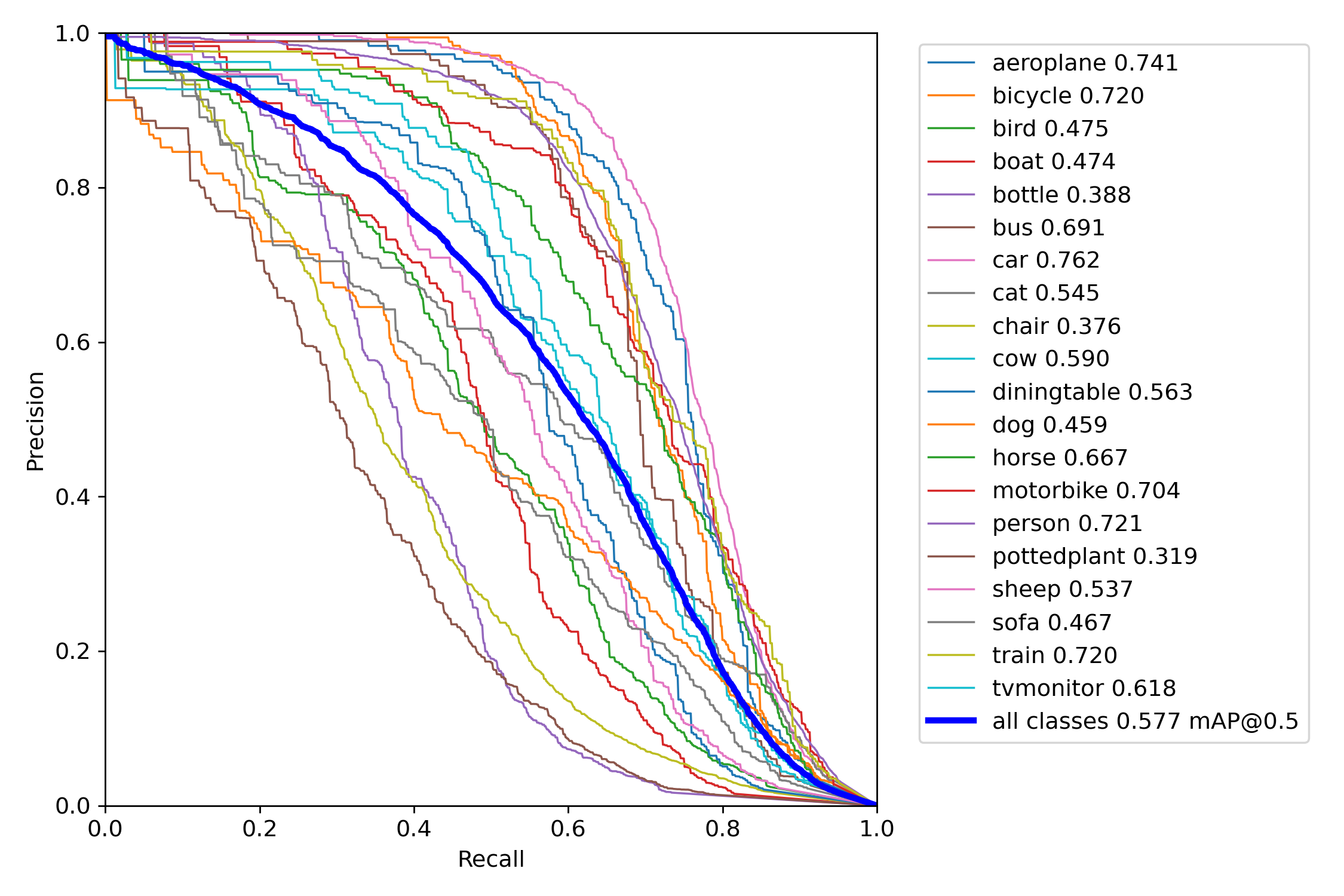}\label{fig_s:2.b}} \\
\subfloat[Underload]{\includegraphics[width=0.45\textwidth]{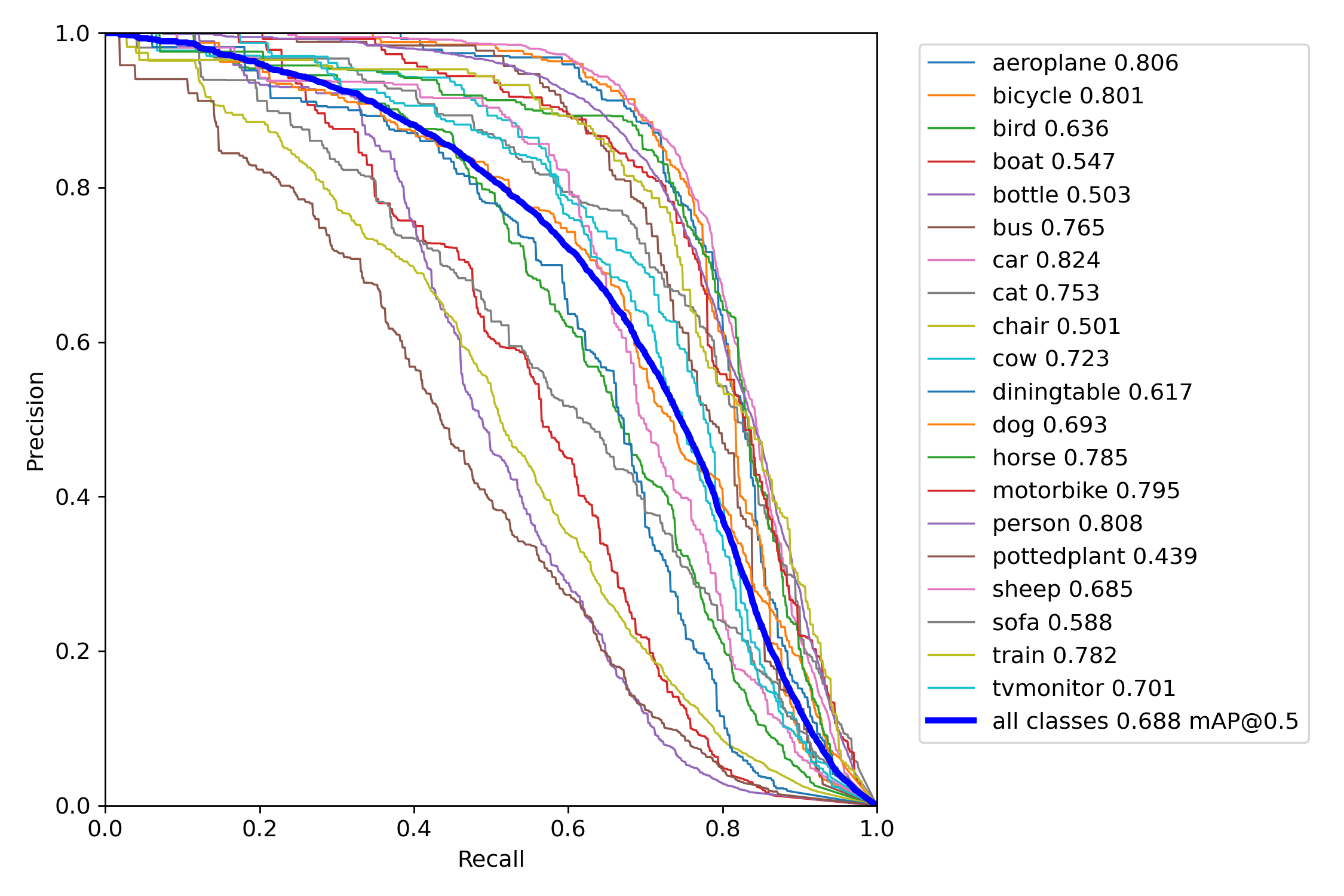}\label{fig_s:2.d}}
\subfloat[OOD]{\includegraphics[width=0.45\textwidth]{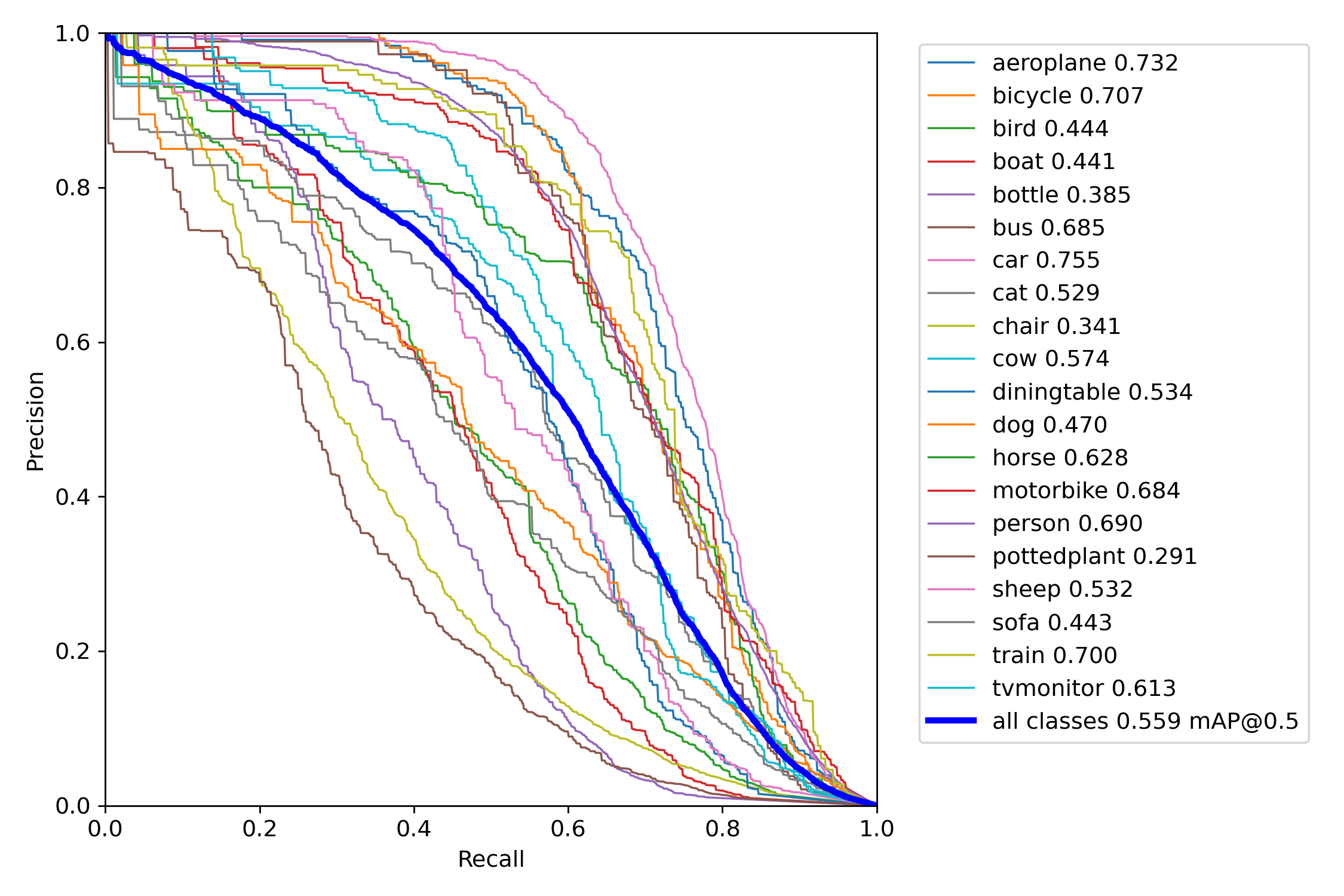}\label{fig_s:2.c}}
\caption{\small{Precision/Recall curves among different AT methods.}}
\label{fig_s_1:pr}
\end{figure*}

As described before, the AT process initiates from the pre-trained model available via the Ultralytics Github site\footnote{https://github.com/ultralytics/yolov5}. We observe the training and validation losses of $\mathcal{L}_\text{box},\mathcal{L}_\text{obj}, \mathcal{L}_\text{cls}$ with the mAP50/95 in Fig.\ref{fig_s_1:training}. We can see that the loss of the proposed Underload is less than the other two AT methods of MTD~\cite{mtd} and OOD~\cite{objadv} on $\mathcal{L}_\text{box}, \mathcal{L}_\text{cls}$ except $\mathcal{L}_\text{obj}$ because Underload employs objectness loss as an adversarial proxy. Injecting adversarial perturbations in the inner optimization against the objectness loss makes the outer minimization more difficult to learn with a larger training loss (but it is below the OOD loss). In the last two columns, the mean average precision of Underload is much higher than both of MTD and OOD from the beginning. The Precision/Recall curves in Fig. \ref{fig_s_1:pr} reveal that the AT methods do not impact any specific category. However, because Underload considers the trade-off between clean and robust accuracy, the accuracy drop is mild and even negligible for some categories (e.g., the accuracy for the bicycle category only drops $0.001$, compared to the drop of $0.082$ and $0.095$ for MTD and OOD). 


\section{Hyperparameter Tuning}
We provide additional experiments of two key hyperparameters: \emph{attack strength}, measured by the $l_2$-norm, and \emph{balance between precision/recall}, measured by the IoU threshold $\Omega_{\text{nms}}$.

\renewcommand{\arraystretch}{1.1} 
\begin{table}[!ht]
\footnotesize
\centering
\begin{tabular}{c c c c c c}
\hline
$l_2$-norm & Attack & Standard & MTD & OOD & \textbf{Underload}* \\ 
\hline
\multirow{3}{*}{10} & \texttt{Daedalus} &  \textbf{\color{red}{73.0}} &  57.6&  55.8&  \color{blue}{68.7}\\
                    & \texttt{Phantom}  &  \textbf{\color{red}{72.1}} &  57.7&  55.7&  \color{blue}{68.7}\\
                    & \texttt{Overload} &  \textbf{\color{red}{71.8}} &  57.7&  55.6&  \color{blue}{68.6}\\
\hline
\multirow{3}{*}{20} & \texttt{Daedalus} &  \textbf{\color{red}{70.9}} &  57.5&  55.6&  \color{blue}{68.4}\\
                    & \texttt{Phantom}  &  \textbf{\color{red}{70.2}} &  57.4&  55.6&  \color{blue}{68.3}\\
                    & \texttt{Overload} &  \color{blue}{66.1} &  57.2&  55.5&  \textbf{\color{red}{68.6}}\\
\hline
\multirow{3}{*}{30} & \texttt{Daedalus} &  \color{blue}{66.6} &  57.2&  55.3&  \textbf{\color{red}{67.9}} \\
                    & \texttt{Phantom}  &  \color{blue}{64.2} &  57.4&  55.8&  \textbf{\color{red}{67.6}} \\
                    & \texttt{Overload} &  51.2&  \color{blue}{57.0}&  55.7&  \textbf{\color{red}{68.4}}  \\
\hline
\multirow{3}{*}{50} & \texttt{Daedalus} &  41.1&  \color{blue}{55.0}&  53.5&  \textbf{\color{red}{62.0}}\\
                    & \texttt{Phantom}  &  32.9&  \color{blue}{56.2}&  54.7&  \textbf{\color{red}{65.4}}\\
                    & \texttt{Overload} &  17.1&  54.0&  \color{blue}{54.8}&  \textbf{\color{red}{59.3}}\\
\hline
\multirow{3}{*}{\textbf{70}} & \texttt{Daedalus} &  12.8&  \color{blue}{50.1}&  48.4&  \textbf{\color{red}{50.3}}\\
                    & \texttt{Phantom}  &  7.5&  \color{blue}{54.7}&  53.3&  \textbf{\color{red}{61.3}}\\
                    & \texttt{Overload} &  4.5&  \color{blue}{49.4}&  49.1&  \textbf{\color{red}{53.3}}\\
\hline
\end{tabular}
\caption{\small{Variation of the attack strengths in terms of $l_2$-norm for YOLOv5s. The \textbf{bolded} $l_2$-norm is the default parameter used in the main text. The {\textbf{\color{red} Best}} and {\color{blue} Second Best} values in each row are marked in {\textbf{\color{red} Red}} and {\color{blue} Blue}. }}
\label{tab:hyp_1}
\end{table}

\subsection{Attack Strength}
To assess attack strength, we select the maximum $l_2$-norm in latency attack~\cite{chen2024overload,wang2021daedalus,ps} for Underload to ensure that our defense remains effective under challenging conditions. As shown in Table \ref{tab:hyp_1}, the $l_2$-norm is set between $[10,70]$. We find that at lower strengths when $l_2$-norm equals to $10$ and $20$, latency attacks have minimal impact on the accuracy of both standard (unprotected) and robust models (Underload AT), since the attack strength is insufficient to push a background region across the boundary margin to generate phantom objects.

However, when $l_2$-norm reaches $30$, it starts to affect the accuracy of the standard model. When the $l_2$-norm exceeds $50$, the accuracy of the unprotected model declines sharply to single digits when $l_2$-norm increases to $70$. On the other hand, we observe that the robust accuracy maintains above the 60\% mAP level for most of the attack strength and is still above 50\% under the maximum $l_2$-norm of $70$. In comparison to MTD and OOD, our approach achieves an accuracy of $0.2$\% to $10.7$\% improvements under different attack strengths. 

\begin{table}
\small
\centering
\begin{tabular}{c c c c c c}
\hline
$\Omega_{\text{nms}}$ & Attack & Standard & MTD & OOD & \textbf{Underload}* \\ 
\hline
\multirow{4}{*}{0.30} & Clean  & \textbf{\color{red}{72.7}} & 58.1 & 57.4 & \color{blue}{68.8} \\
                    & \texttt{Daedalus} & 12.7 & \textbf{\color{red}{50.4}} & 49.5 & \color{blue}{49.9} \\
                    & \texttt{Phantom}  & 7.2 & \color{blue}{54.8} & 54.7 & \textbf{\color{red}{61.2}} \\
                    & \texttt{Overload} & 4.1 & 49.1 & \color{blue}{50.3} & \textbf{\color{red}{52.6}} \\
\hline
\multirow{4}{*}{0.45} & Clean  & \textbf{\color{red}{73.6}} & 58.7 & 57.8 & \color{blue}{69.5} \\
                    & \texttt{Daedalus} & 13.0 & \textbf{\color{red}{50.9}} & 50.0 & \color{blue}{50.5} \\
                    & \texttt{Phantom}  & 7.4 & \color{blue}{55.6} & 55.1 & \textbf{\color{red}{61.9}} \\
                    & \texttt{Overload} & 4.3 & 49.7 & \color{blue}{50.6} & \textbf{\color{red}{53.4}} \\
\hline
\multirow{4}{*}{\textbf{0.60}} & Clean & \textbf{\color{red}{73.3}} & 57.7 & 55.9 & 68.9 \\
                    & \texttt{Daedalus} & 12.8 &  \color{blue}{50.1} &  48.4&  \textbf{\color{red}{50.3}} \\
                    & \texttt{Phantom}  & 7.5 &  \color{blue}{54.7} &  53.3 & \textbf{\color{red}{61.3}} \\
                    & \texttt{Overload} & 4.5 &  \color{blue}{49.4} &  49.1 & \textbf{\color{red}{53.3}} \\
\hline
\multirow{4}{*}{0.75} & Clean  & \textbf{\color{red}{71.4}} & 53.7 & 50.7 & \color{blue}{65.7} \\
                    & \texttt{Daedalus} & 12.1 & \color{blue}{46.6} & 43,7 & \textbf{\color{red}{47.5}} \\
                    & \texttt{Phantom}  & 7.4 & \color{blue}{51.1} & 48.5 & \textbf{\color{red}{58.6}} \\
                    & \texttt{Overload} & 4.4 & \color{blue}{46.6} & 45.1 & \textbf{\color{red}{51.3}} \\
\hline
\multirow{4}{*}{0.9} & Clean   & \textbf{\color{red}{62.5}} & 39.6 & 35.1 & 52.7 \\
                    & \texttt{Daedalus} & 12.0 & \color{blue}{33.9} & 29.5 & \textbf{\color{red}{36.9}} \\
                    & \texttt{Phantom}  & 6.3 & \color{blue}{37.6} & 33.7 & \textbf{\color{red}{46.5}} \\
                    & \texttt{Overload} & 3.7 & \color{blue}{35.0} & 31.6 & \textbf{\color{red}{41.1}} \\
\hline
\end{tabular}
\caption{\small{Variations of the IoU threshold $\Omega_{\text{nms}}$ in YOLOv5s. The \textbf{bolded} $\Omega_{\text{nms}}$ is the default parameter used in the main text. The {\textbf{\color{red} Best}} and {\color{blue} Second Best} values in each row are marked in {\textbf{\color{red} Red}} and {\color{blue} Blue}. The first row of ``Clean'' compares the clean accuracy drop with different AT methods under different $\Omega_{\text{nms}}$. }}
\label{tab:hyp_2}
\end{table}

\subsection{Balance between Precision/Recall}

The Intersection over Union (IoU) threshold is an important parameter that balances the precision and recall in object detection. A higher IoU threshold during the NMS process retains fewer candidate boxes, which reduces the occurrence of false positives and enhances the model's precision. However, setting the IoU threshold too high may inadvertently remove some true positives, thereby reducing the recall. Conversely, reducing the IoU threshold can enhance recall by keeping more candidate boxes, but it also leads to an increase in overlapping detection, which ultimately decreases the precision. 

Under latency attacks, the IoU threshold can be adjusted to simulate various attack scenarios. Setting the IoU threshold to $1.0$ keeps all the candidate boxes (no box is removed), thereby retaining the artifacts produced by the latency attack that emulates the worst-case scenario. 
We evaluate five IoU thresholds ranging from 0.3 to 0.9, with an increment of $0.15$, covering a wide range of IoU threshold values. We observe that both the clean and robust accuracy of the standard and robust models exhibit an initial increase followed by a decrease as the IoU threshold increases. Among the selected IoU thresholds, the highest accuracy occurs at an IoU threshold of $0.45$. At this threshold, the reduction of Underload in clean accuracy is minimal, with a decrease of only $4.1$\%. In most cases, the robust accuracy of the Underload outperforms the other two AT methods. However, there are a few outliers in the low IoU cases (when the IoU threshold is $0.3$ or $0.45$), the robust accuracy of the MTD exceeds that of the Underload. We conjecture that it is due to the Daedalus method, which simultaneously optimizes the confidence and size of the phantom objects, may generate some high-confidence and large-area phantoms (compared with other latency attacks). When the IoU threshold is low, these phantoms can interfere with natural objects, leading to a reduction in the robust accuracy. 

\section{Framework and Hardware Optimization under Latency Attacks}

\begin{table*}[t]
\small
\centering
\begin{tabular}{ccccccc}
\hline
\multirow{2}{*}{Model} & \multirow{2}{*}{Device} & \multirow{2}{*}{Attack} & \multicolumn{2}{c}{Standard} & \multicolumn{2}{c}{\textbf{Underload}*} \\
\cmidrule(lr){4-5} \cmidrule{6-7}
        &    &   &  ONNX        & TensorRT       & ONNX        & TensorRT        \\
\hline
\multirow{8}{*}{YOLOv5s}&\multirow{2}{*}{1650Ti Laptop}& Clean & 19.4 &  14.1  & 19.0 & 14.2 \\
                                        &    &\texttt{Overload}& 69.2 &  64.9  & 18.7 & 14.0 \\
\cmidrule(lr){2-7}
&\multirow{2}{*}{4070Ti Super}  & Clean  &  7.4  &  6.3  &  7.2   &  6.2   \\
&                      &\texttt{Overload}&  31.1 &  28.6  &  7.0   &  6.3  \\
\cmidrule(lr){2-7}
&\multirow{2}{*}{Jetson Orin NX}& Clean & 30.3 & 19.2 & 30.1  & 20.0 \\
            & &\texttt{Overload} & 100.8 & 87.7 & 29.8  & 19.9 \\
\cmidrule(lr){2-7}
&\multirow{2}{*}{Jetson Xavier NX}& Clean & 57.7 & 30.8 &  57.6  &  30.0  \\
                     & &\texttt{Overload} & 447.2 & 418.5 & 56.0 &  31.0  \\
\hline
\multirow{8}{*}{YOLOv8s}&\multirow{2}{*}{1650Ti Laptop}& Clean & 23.1 &  13.8  & 23.0 & 14.0 \\
                                        &    &\texttt{Overload}& 25.3 &  16.0  & 23.1 & 13.9 \\
\cmidrule(lr){2-7}
&\multirow{2}{*}{4070Ti Super}  & Clean  &  8.3  &  2.9  &   8.4   &  3.0    \\
&                      &\texttt{Overload}& 9.4   &  4.0  &   8.3   &  2.8  \\
\cmidrule(lr){2-7}
&\multirow{2}{*}{Jetson Orin NX}& Clean & 30.1 & 18.2 & 30.0  & 18.0 \\
            & &\texttt{Overload} & 33.2 & 21.7 & 30.3  & 18.1 \\
\cmidrule(lr){2-7}
&\multirow{2}{*}{Jetson Xavier NX}& Clean & 58.1 & 42.2 &  58.0 & 40.0 \\
                     & &\texttt{Overload} & 59.8 & 45.7 &  57.6 & 40.3 \\
\hline
\end{tabular}
\caption{\small{Different frameworks of YOLOv5s and YOLOv8s model inference time (ms) in FP32. }}
\label{tab:onnx_inf}
\end{table*}

\begin{table}[t]
\centering
\begin{tabular}{c c c}
\hline
Package & Desktop Ver. & Edge Device Ver. \\ 
\hline
CUDA & 11.7 & 11.4 \\
Ultralytics & 8.2.100 & 8.2.100 \\
ONNX & 1.14.0 & 1.17.0 \\
ONNXRuntime & 1.16.0 & 1.16.0 \\
TensorRT & 8.6.0 & 8.5.2 \\
\hline
\end{tabular}
\caption{\small{ONNX and TensorRT models inference environments.}}
\label{tab:onnx_env}
\end{table}

In addition to PyTorch implementation, we also convert YOLOv5s and YOLOv8s to other implementations including ONNX and TensorRT for specialized acceleration. For reproducing purposes, the environment setup is shown in the Table~\ref{tab:onnx_env}. We employ the same method to attack the implementations of ONNX and TensorRT. In details, we export ONNX and TensorRT models using the official implementation, which convert only the backbone without utilizing the \texttt{INMSLayer} or \texttt{EfficientNMSPlugin}. Other configurations remain the same as~\cref{sec:evaluation}.  

From Table~\ref{tab:onnx_inf}, we find that latency attacks affect models across different implementations even for TensorRT. Despite of hardware-specific optimizations, the execution time still increases by $1.1-13.5\times$ under the \texttt{Overload} attack. In TensorRT, \texttt{EfficientNMSPlugin} is also vulnerable because as per to our evaluation, its execution time increases with the number of candidate boxes. Fortunately, the AT models exported from our \texttt{Underload} defense are able to defend against the corresponding latency attacks and portable to different frameworks and edge devices.  

\end{document}